\documentclass{article}

\usepackage{arxiv}
\usepackage[utf8]{inputenc}
\usepackage[T1]{fontenc}
\usepackage{amsfonts}
\usepackage{microtype}
\usepackage{nicefrac}
\newlength{\extralength}
\usepackage{amsmath}
\usepackage{amsthm}
\usepackage{amsbsy}
\usepackage{tikz}
\usepackage{booktabs}
\usepackage{tabularx}
\usepackage{array}
\usepackage{multirow}
\usepackage{chngcntr}

\usepackage{graphicx}
\usepackage{float}
\usepackage{subcaption}
\usepackage{wrapfig}
\usepackage{changepage}

\usepackage{xcolor}
\usepackage{pifont}

\usepackage{algorithm}
\usepackage{algpseudocode}

\usepackage[numbers,square]{natbib}
\usepackage{url}
\usepackage{hyperref}

\newtheorem{definition}{Definition}[section]

\newtheorem{example}{Example}[section]

\title{Nested Ensemble Learning with Topological Data Analysis for Graph Classification and Regression}

\author{
    Innocent Abaa\\
    Department of Mathematics\\
    Juniata College\\
    \texttt{iga100@juniata.edu} \\
        \And
   Umar Islambekov\\
    Department of Mathematics and Statistics\\
    Bowling Green State University\\
    \texttt{iumar@bgsu.edu} \\
}

\begin{document}
\maketitle

\vspace{-0.5cm}

\begin{abstract} We propose a nested ensemble learning framework that utilizes Topological Data Analysis (TDA) to extract and integrate topological features from graph data, with the goal of improving performance on classification and regression tasks. Our approach computes persistence diagrams (PDs) using lower-star filtrations induced by three filter functions: closeness, betweenness, and degree2 centrality. To overcome the limitation of relying on a single filter, these PDs are integrated through a data-driven, three-level architecture. At Level-0, diverse base models are independently trained on the topological features extracted for each filter function. At Level-1, a meta-learner combines the predictions of these base models for each filter to form filter-specific ensembles. Finally, at Level-2, a meta-learner integrates the outputs of these filter-specific ensembles to produce the final prediction. We evaluate our method on both simulated and real-world graph datasets. Experimental results demonstrate that our framework consistently outperforms base models and standard stacking methods, achieving higher classification accuracy and lower regression error. It also surpasses existing state-of-the-art approaches, ranking among the top three models across all benchmarks.
\end{abstract}



\section{Introduction}

Graphs serve as natural representations of complex systems in a wide range of domains, including social networks, biological systems, and infrastructure networks. With the increasing prevalence of graph-structured data, there is a growing demand for models that can effectively capture both local and global structural patterns to support predictive tasks such as classification and regression. Ensemble learning has emerged as a particularly powerful strategy in this context. By combining multiple base models, ensemble methods enhance predictive accuracy, reduce overfitting, and improve generalization. Among these techniques, stacked generalization (or stacking)~\cite{wolpert1992stacked} is notable for its ability to integrate diverse base learners through a higher-level meta-learner. This architecture allows the model to leverage complementary patterns learned by each base learner, making it well-suited for heterogeneous and structurally complex datasets such as graphs. The theoretical advantage of stacking lies in its ability to balance the bias--variance trade-off, which is particularly crucial in domains characterized by irregular, non-Euclidean structures.

{A key avenue for enhancing stacking-based approaches on graph data is the incorporation of domain-specific structural features that reflect the underlying topology of the graph. In this regard, Topological Data Analysis (TDA) provides a principled, mathematically grounded framework for these purposes. Rooted in algebraic topology and computational geometry, TDA offers tools to study the shape structure underlying} data~\cite{carlsson2009topology}. The most widely used tool in TDA is \textit{persistent homology}which tracks how topological features (such as connected components, loops, and voids) appear and disappear as a graph is analyzed at different scales. This is achieved by building a sequence of nested topological spaces, known as a filtration, on top of the data and recording the birth and death of topological features as the filtration progresses. The evolution of these features is summarized in a persistence diagram, where each point represents a feature, with its coordinates indicating the scale at which it was ``born'' and ``died.'' This provides a compact summary of the data's shape across multiple scales. Unlike traditional graph-theoretic metrics, which are limited to local connections, persistent homology captures both local and global patterns. This makes it a robust method for analyzing graph structure, as it is less sensitive to noise and small changes in the data.

A central step in applying TDA is choosing an appropriate \emph{filtration}, {a nested sequence of simplicial complexes (collections of vertices, edges, and higher-dimensional geometric shapes called simplices) used to track topological features. Commonly used filtrations, such as the Vietoris--Rips and Čech, are designed for point clouds in a metric space, where simplices are added based on pairwise distances. However, this approach is problematic for graph-structured data for several reasons. While it is possible to embed a graph into a metric} space~\cite{hajij2017persistent,you2022comparingmultiplelatentspace}, doing so often leads to a potential loss of crucial node connectivity information. Furthermore, applying metric-based filtrations to the resulting point cloud becomes computationally expensive for large graphs, as the Vietoris--Rips complex can grow exponentially with the number of nodes. For graph data, an effective alternative is the lower-star filtration~\cite{hajij2021persistenthomologygraphsrepresentation}, which constructs a valid filtration using a scalar-valued function defined on the graph's nodes or edges. This method circumvents the need for an explicit metric embedding by leveraging node-level attributes like degree, centrality, or spectral features to dictate the order in which simplices are added to the filtration. A simplex enters the filtration at the maximum function value of its vertices, satisfying a sublevel set condition that ensures the filtration is nested and mathematically well-defined for persistent homology. By utilizing node-level functions rather than pairwise distances, the lower-star filtration respects the graph's intrinsic structure and seamlessly integrates both topological and attribute-based information, avoiding the computational and information-loss issues associated with metric embeddings~\cite{islambekov2023fasttopologicalapproachpredicting}.

{While the lower-star filtration offers a flexible and mathematically valid framework for analyzing graph-structured data based on node or edge attributes, a significant challenge lies in selecting an appropriate scalar function to assign filtration values, as its choice critically influences the resulting topological summaries. Common options include node attributes, such as degree }~\cite{niepert2016learningconvolutionalneuralnetworks},  spectral embeddings~\cite{zhao2019learningmetricspersistencebasedsummaries}, or domain-specific features~\cite{carriere2020perslay, hofer2018deeplearningtopologicalsignatures}, which {critically influence the resulting topological summaries.  A poorly chosen function can obscure relevant topological signals or introduce spurious features, potentially degrading model performance. Since no single function is universally optimal, relying on a fixed, manually selected filter, which is common in many existing TDA approaches}~\cite{niepert2016learningconvolutionalneuralnetworks,carriere2020perslay, hofer2018deeplearningtopologicalsignatures, zhao2019learningmetricspersistencebasedsummaries}, can restrict a model's flexibility and expressive power, particularly when that function fails to highlight the topological patterns most relevant to the learning task.

To move beyond the limitations of single, hand-crafted filters, recent research has explored several advanced strategies. One approach is the use of multi-filter filtrations, where diverse filtrations are constructed from multiple structural perspectives, and their persistence diagrams are vectorized and concatenated into a single feature vector for downstream tasks~\cite{agerberg2025certifying, obray2021filtration}. A complementary, more sophisticated direction involves learning the filter functions directly from data, embedding their design into the learning process so that filtrations are adaptively determined rather than fixed in advance. For example, frameworks for point clouds adaptively learn filtrations via neural networks~\cite{nishikawa2023adaptive} or through topological autoencoders that use a persistent homology loss to preserve structure~\cite{cornell2021usingtopologicalautoencodersfiltering}. For graphs, frameworks like TOGL extend this idea by learning topology-aware representations end-to-end within a graph neural network, removing the reliance on fixed descriptors~\cite{horn2022topologicalgraphneuralnetworks}. A third, complementary but more complex strategy is multiparameter persistent homology (MPH), which applies filtrations across multiple independent parameters \mbox{simultaneously~\cite{carriere2020perslay, bu2023interplay, hofer2018deeplearningtopologicalsignatures}}. {While MPH captures a richer mathematical object that describes the interactions between different filtration parameters, it also introduces significant computational and interpretational challenges. Despite recent efforts to create vectorized summaries that make MPH more accessible}~\cite{NEURIPS2023_d75c474b}, {these trade-offs highlight the continued relevance of alternative multi-filter strategies.}

Building on these advances, we propose a model-based, data-driven ensemble learning framework that integrates multiple filter functions, with their contributions adaptively optimized through a principled, two-level meta-learning process. Unlike previous methods that either treat each filter independently or simply concatenate their features, our approach first computes persistence diagrams for multiple filter functions using lower-star filtrations. Each filter function, defined by a specific node attribute, guides the construction of the simplicial complex. These persistence diagrams are then vectorized into distinct topological feature sets, which serve as the foundation for our model building.

The key innovation of our framework lies in how we combine these multiple filter functions in a fully data-driven manner. The process involves a three-level architecture:

\begin{enumerate}
    \item \textbf{Base Model Training (Level-0):}  
{For each filter function, a diverse set of base machine learning models is trained independently. This parallel training strategy enables each model to specialize in identifying the unique predictive patterns present within the topological features extracted by its corresponding filter.}

 \item \textbf{Filter-specific Ensemble (Level-1):}
hl{For each filter function, a first-level meta-learner is trained to optimally combine the predictions of that filter's diverse base models from Level-0. This process results in a single, refined, stacked ensemble prediction for each filter function.}
    
\item \textbf{Cross-filter Ensemble (Level-2):} {A second-level meta-learner is trained on the outputs from all filter-specific ensembles from Level-1. This meta-learner learns to adaptively assign importance weights to each filter, thereby empirically determining its relative influence on the final nested ensemble prediction.}

\end{enumerate}

This hierarchical weighting mechanism, grounded in empirical evidence, directly learns which filters are most informative for the task. It eliminates the need for manual filter selection, as even randomly chosen or unconventional filters can be included and their relevance automatically assessed. The model dynamically prioritizes filters that contribute most to predictive performance while down-weighting less useful ones. By unifying multiple topological perspectives through this automated ensemble structure, our method consistently enhances performance on downstream tasks such as classification \linebreak  and regression.

 \textls[5]{We validate our approach on both synthetic and real-world data, including several widely used benchmark datasets from the TUDataset collection~\cite{kersting2016benchmark}: \textit{IMDB-BINARY}, \textit{IMDB-MULTI}, \textit{REDDIT-BINARY}, and \textit{PROTEINS}. The results demonstrate that our proposed nested ensemble framework consistently outperforms both the individual base models and standard stacking methods across classification and regression tasks. Moreover, topological features derived from persistent homology show superior performance compared to traditional graph-theoretic descriptors, underscoring the benefits of integrating TDA with ensemble learning for graph-based machine learning. On the TUDataset benchmarks, we further demonstrate the effectiveness of our method by comparing it to the top five existing topological approaches. Our preliminary results are promising and indicate that the information extracted by our framework can be effectively utilized for graph classification~tasks.}

The remainder of this paper is organized as follows. Section~\ref{sec:background} introduces the theoretical foundations of persistent homology and describes how we extract topological features from graphs. Section~\ref{sec:methodology} details the proposed nested ensemble learning framework. \mbox{Section~\ref{sec:results}} presents our experimental results, and Section~\ref{sec:conclusion} provides a summary and discusses directions for future research.

\section{Background and Preliminaries}
\label{sec:background}

In this section, we introduce the foundational concepts of topological data analysis (TDA), 
focusing on simplicial complexes, filtrations, homology, and persistent homology, 
following the notation and conventions in {~\cite{edelsbrunner2010computational,ghrist2008barcodes,otter2017roadmap}}. 
These concepts form the theoretical basis for the methodology developed in this study.

\subsection{TDA Background Theory}

Topological Data Analysis (TDA) provides a mathematical framework for uncovering the intrinsic geometric and topological structures underlying data. A central tool in TDA is \emph{persistent homology}, which tracks the evolution of topological features (e.g., connected components, loops, and voids) across multiple scales. The resulting output is often summarized in the form of a \emph{persistence diagram} (PD), a multiset of points (A multiset of points means that different topological features can have the same birth and death times in the persistence diagram.)
 $(b,d) \in \mathbb{R}^2$, where each point corresponds to a topological feature that appears (is ``born'') at scale $b$ and disappears (or ``dies'') at scale $d$ (see Figure \ref{fig:placeholder}). These diagrams serve as topological signatures of the data, capturing information such as connected components, loops, and voids across multiple scale values.

{To compute a persistence diagram, we must first equip the data (often given as a point cloud, image, or graph) with a topological structure through the construction of a \textit{simplicial complex}. A simplicial complex provides a combinatorial framework that captures relationships between data points. This is accomplished through various constructions, such as the Vietoris--Rips, Čech, or lower star filtration, which approximate the underlying topological structure of the space from which the data were sampled. This process allows for the systematic study of topological features using tools from algebraic topology, such as homology and persistent homology.}

\begin{figure}[H]
    \includegraphics[width=1.0\textwidth]{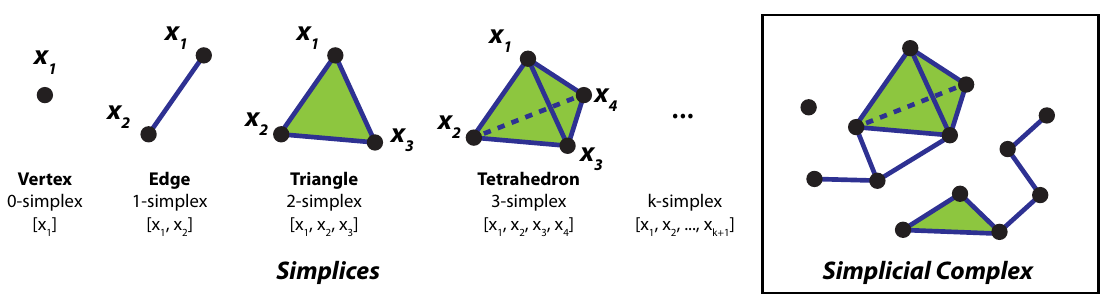}
    \caption{Geometric
 representation of simplices of increasing dimension-vertices (0-simplices), edges (1-simplices), triangles (2-simplices), and a tetrahedron (3-simplex)-shown on the left. On the right, these building blocks are combined to form a simplicial complex, illustrating how simplices connect through shared faces. Image reproduced from Zhang et al.~\cite{zhang2020topological}.}
    \label{tab:simplices}
\end{figure}

{An \emph{(abstract) simplicial complex} $ K $ is a collection of finite subsets of a vertex set $ V $}, 
satisfying the following two properties. First, every vertex $ v \in V $ must be included as a singleton set $ \{v\} \in K $. Second, the complex must be closed under taking subsets, i.e., if $ \tau \in K $, then any subset $ \sigma \subseteq \tau $ is also in $ K $. In this context, any subset $\tau \in K$ containing $p+1$ vertices is called a $p$-simplex and has dimension $p$, denoted $\dim(\tau) = p$. A subset $\sigma \subseteq \tau$ is called a \emph{face} of $\tau$, written $\sigma \preceq \tau$. The set of all simplices in $K$ of dimension $d$ is denoted by $K_d$, and the overall dimension of $K$ is the highest dimension among its simplices. {There is a geometric realization of simplicial complexes in a suitable Euclidean space,} in which a vertex corresponds to a 0-simplex, an edge to a 1-simplex, a filled triangle to a 2-simplex, and a tetrahedron to a 3-simplex, and so on (see Figure~\ref{tab:simplices}); in this realization, different simplices intersect only along their common faces. Example
 \ref{ex:lowerstar1} illustrates both the concept of a simplicial complex and the process by which it is constructed.

\begin{figure}[H]
\centering
    \includegraphics[width=0.25\linewidth]{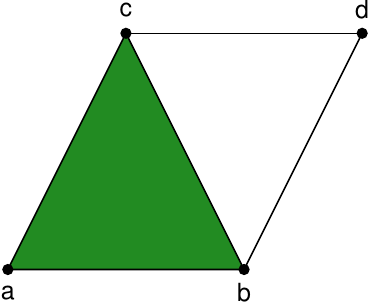}
    \caption{Simplicial complex $X$.}
    \label{fig:enter}
\end{figure}

\begin{example}
\label{ex:lowerstar1} 
Considerthe simplicial complex $X$ shown in Figure~\ref{fig:enter}, which has the vertex set 
$V = \{a, b, c, d\}$ and the collection of simplices $K = \{\{a,b,c\}, \{a,b\}, \{a,c\}, \{b,c\}$, $\{b,d\}, \{c,d\}$, $\{a\}, \{b\}, \{c\}, \{d\}\}.$
The $0$-simplices, denoted $K_0 = \{\{a\}, \{b\}, \{c\}, \{d\}\}$, are  simply the vertices themselves.
The $1$-simplices, $K_1 = \{\{a,b\}, \{a,c\}, \{b,c\}, \{b,d\}, \{c,d\}\}$, represent the edges connecting these vertices.
Finally, the $2$-simplex $K_2 = \{\{a,b,c\}\}$ corresponds to the filled triangle formed by vertices $a$, $b$, and $c$.
\end{example}

\vspace{-15pt}

A key step in TDA involves constructing a filtration-a nested sequence of simplicial complexes that progressively captures the evolving topological features of the data at multiple scales or thresholds. {One common approach to building filtrations is through the sublevel sets of a monotone function defined on the simplices of a simplicial complex.} {Specifically, let $K$ be a simplicial complex, and let $g:K\rightarrow \mathbb{R}$ be a real-valued function} defined on its simplices. {The function $g$ is said to be monotone if, for every pair of simplices} $\sigma \subseteq \tau \in K$, it satisfies $g(\sigma )\le g(\tau )$. This condition ensures that for each threshold $t\in \mathbb{R}$, the sublevel set $K_{t}=\{\sigma \in K\mid g(\sigma )\le t\}$ forms a valid simplicial subcomplex of $K$, because if a simplex is in $K_{t}$, its faces are also guaranteed to be in $K_{t}$ by the monotonicity property. Let $\{c_{1}<c_{2}<\cdots <c_{N}\}$ denote the ordered set of distinct function values taken by $g$. The sublevel set filtration induced by $g$ is then given by the nested sequence:
\begin{equation}
\emptyset = K_1(g) \subseteq K_2(g) \subseteq \cdots \subseteq K_N(g) = K
\end{equation}
where each $K_{i}(g)=\{\sigma \in K\mid g(\sigma )\le c_{i}\}$. The number of filtration steps $N$ corresponds to the number of unique function values attained by $g$ over the simplices in $K$.

\subsubsection{Homology and Persistent Homology}
\label{section2.1.1}

To understand how topological features such as connected components, loops, and voids evolve across a filtration, we need algebraic tools that can precisely capture these structures. This is exactly the role of \emph{homology}, which formalizes these features using algebraic objects called homology groups.

{Given a sequence of nested simplicial complexes-called a filtration-we can apply homology to each complex to systematically identify and track topological features across} {multiple scales. The technique of \emph{persistent homology} tracks how these features, such as} connected components ($H_{0}$), loops  ($H_{1}$), and voids  ($H_{2}$), evolve and persist over the course of the filtration. At each stage, a simplicial complex $K$ is constructed from a set of vertices, edges, triangles, and possibly higher-dimensional simplices. To uncover the topological information contained in these complexes, we associate to them algebraic structures called \emph{chain groups}. For each dimension $d$, the chain group $C_d(K)$ is defined as the vector space (or free module) generated by the $d$-simplices in $K$, {with coefficients chosen in a fixed} {field
($ \text{like } \mathbb{Z}_2 \text{ or } \mathbb{R} $).} An element of $C_d(K)$ is called a \emph{$d$-chain}. It is a linear combination of $d$-simplices, written as
\begin{equation}
c = a_1 \sigma_1 + a_2 \sigma_2 + \cdots + a_m \sigma_m,
\end{equation}
where the $\sigma_i$ are $d$-simplices in $K$ and the $a_i$ are coefficients from the chosen field. Next, we connect different dimensions using \emph{boundary operators} 
$\partial_d: C_d(K) \to C_{d-1}(K)$.
These operators map each $d$-simplex to a sum of its $(d-1)$-dimensional faces.
For example, for a simplex $[v_0, v_1, \dots, v_d]$,
\begin{equation}
\partial_d([v_0, v_1, \dots, v_d])
= \sum_{j=0}^d (-1)^j [v_0, \dots, \hat{v}_j, \dots, v_d],
\end{equation}
where $\hat{v}_j$ means that the vertex $v_j$ is omitted.
When working over $\mathbb{Z}_2$, we can ignore the signs and simply track whether faces are present or absent.

A key property of the boundary operator is that applying it twice yields zero, i.e., $\partial _{d-1}\circ \partial _{d}=0.$ {We define a $d$-cycle as a $d$-chain whose boundary is zero, i.e., it is in the kernel of} the boundary operator $\partial _{d}$. This space is denoted by $Z_{d}(K)=\text{ker}(\partial _{d})$. Similarly, a $d$-boundary is a $d$-chain that is the boundary of some $(d+1)$-chain. These form the image of the boundary operator $\partial _{d+1}$, denoted by $B_{d}(K)=\text{im}(\partial _{d+1})$. Because applying the boundary operator twice yields zero ($\partial _{d-1}\circ \partial _{d}=0$), every boundary is automatically a cycle. This means we have the following subgroup relationship: $B_{d}(K)\subseteq Z_{d}(K).$

The $d$-th homology group is then defined as the quotient:
\begin{equation}
H_d(K) = Z_d(K)/ B_d(K),
\end{equation}
which captures cycles that are not boundaries. The dimension of $H_{d}(K)$, called the $d$-th Betti number, $\beta _{d}$, counts the number of independent $d$-dimensional holes. It is calculated as $\beta _{d}=\dim (Z_{d}(K))-\dim (B_{d}(K))$. For $d\ge 1$, $\beta _{d}$ is interpreted as counting the number of independent loops ($\beta _{1}$) and higher-dimensional voids ($\beta _{d}$ for $d>1$). {In ordinary homology,} $\beta _{0}$ {counts the number of connected components.} {However, in some contexts, particularly for single-point spaces, reduced homology is used, where the zeroth reduced Betti number is one less than the number of connected components}~\cite{wikipedia:reduced_homology}.

Persistent homology extends homology to filtrations, allowing us to track how topological features appear and disappear across multiple scales. As we build the simplicial complex in the filtration, topological features are both born and destroyed. {For example, new connected components emerge (increasing the Betti number} $\beta _{0}$) and can later merge with other components (decreasing $\beta _{0}$). {Similarly, new loops can be created by the addition of edges (increasing} $\beta _{1}$) and are destroyed when a triangle fills them in (decreasing $\beta _{1}$). This dynamic process is captured in a persistence diagram, which records the birth and death of each topological feature, providing a concise summary of its lifespan across the filtration. Given a filtration
\begin{equation}
\emptyset = K_0 \subseteq K_1 \subseteq \cdots \subseteq K_N = K,
\end{equation}
the inclusion maps $ K_a \hookrightarrow K_b $ induce homomorphisms between homology groups:
\begin{equation}
\iota_d^{a \to b}: H_d(K_a) \to H_d(K_b).
\end{equation}
Each homology class can be associated with a birth time, $b_{x}$, indicating when it first appears in the filtration, and a death time, $d_{x}$, indicating when it either becomes trivial or merges into an older feature. Collecting these persistence intervals, $[b_{x},d_{x})$, across all dimensions yields the persistence diagram for a given homological degree $d$, which provides a concise summary of the multiscale topological features present in the data.

Although persistent homology is a highly general framework that can be applied to filtrations of any topological space, in this work, we focus on filtrations constructed from scalar functions defined on the nodes of a graph. A simple and widely adopted method for building such filtrations is the \emph{lower-star filtration}, which constructs nested subcomplexes based on real-valued functions assigned to the vertices. This approach is especially advantageous when distance metrics are not available or do not make sense, or when meaningful scalar attributes-such as structural properties or centrality scores-can be derived from the graph and assigned to its nodes.

Let $ G = (V, E; g) $ be an undirected graph, where $ V $ is the set of nodes, $ {E} \subseteq \{\{u, v\} \mid u, v \in V,\, u \neq v \} $ is the set of edges, and $ g: V \to \mathbb{R} $ assigns real-valued attributes to the nodes. This graph can be naturally interpreted as a one-dimensional simplicial complex; the nodes are $0$-simplices and edges are $1$-simplices. To capture more complex topological structures beyond pairwise connections, we extend this to higher-dimensional simplices. For instance, a triangle formed by three mutually connected nodes corresponds to a $2$-simplex, and a tetrahedron formed by four fully connected nodes corresponds to a $3$-simplex, and so forth. The resulting complex, often called a \textit{clique complex} or \textit{flag complex}, encodes all complete subgraphs of the original graph as simplices. For notational simplicity, we continue to denote this extended simplicial complex by $ G $.

To analyze the topology of $G$, we build a filtration by extending the function $g$ from the vertices to all simplices using the \emph{maximum rule}. Specifically, for any simplex $\tau \subseteq G$ (viewed as a simplicial complex), we define $g(\tau) = \max \{ g(v) : v \in \tau \}.$
This extension ensures monotonicity; if $\sigma \subseteq \tau$, then $g(\sigma) \leq g(\tau)$. As a result, the sublevel sets of $g$ form valid simplicial complexes, which allows us to define a filtration. The resulting \textit{Lower-Star filtration} is defined as a sequence of nested subcomplexes $
G_t = \{ \tau \subseteq G \mid g(\tau) \leq t \},$
where the filtration parameter $ t $ increases over the range of values taken by $ g $. In this construction, a simplex enters the filtration at the smallest threshold $t$ such that all its vertices satisfy $g(v) \leq t$. Because the value on a simplex depends on the maximum among its vertices, this method provides a consistent, hierarchical way to explore the topological features of the~graph.

In the Lower-Star filtration, the choice of function values assigned to the vertices is critical, as it directly determines how the filtration unfolds and what topological features are captured. These function values act as filtration parameters, deciding the exact stage at which each simplex enters the filtration, ultimately shaping the topological summary we obtain. In this work, we define the vertex function $g$ based on three types of node attributes: closeness centrality, betweenness centrality, and a custom measure we call degree2 centrality. {Closeness centrality measures how close a node is, on average, to all other nodes in the network, based on the shortest-path distances. When analyzing unweighted graphs, each edge is treated as having unit length, and distances between nodes are computed as the number of edges along the shortest path. This node-level measure reflects the overall accessibility or influence of a node within the graph.} Betweenness centrality measures how often a node lies on the shortest paths between other nodes, highlighting nodes that serve as bridges or bottlenecks within the graph. The degree2 centrality metric is defined as the count of unique nodes connected to a given node through paths of length one or two. By incorporating nodes at a distance of two hops, this measure provides a more extensive characterization of a node's local neighborhood structure than that offered by standard degree centrality.

Each of these filter functions induces a distinct ordering of simplices, which governs how connected components, cycles, and other topological features appear and persist across scales. Applying persistent homology to these filtrations yields a persistence diagram that records the birth and death of each topological feature as the threshold \textit{t} varies. These diagrams provide compact, multi-scale summaries of the graph’s topological structure relative to the chosen function $ g $. (See Example \ref{ex:lower} for an illustration of how lower-star filtrations can be used to compute persistence diagrams.)

\begin{figure}[H]
\centering
    \includegraphics[width=0.3\linewidth]{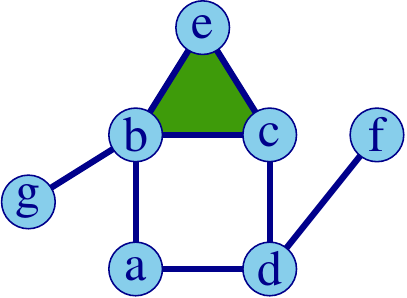}
    \caption{A simple house graph $ G $ consisting of seven nodes and eight edges.
}
    \label{fig:enter-label}
\end{figure}

\begin{example}
\label{ex:lower}
{Consider a graph $ G = (V, E) $ with vertex set} $ V = \{a, b, c, d, e, f,g\} $  
and edge set $ E = \{(a,b), (b,c), (c,d), (d,a), (b,e), (c,e), (d,f),(b,g)\} $. {This graph can be viewed as a 1-dimensional simplicial complex with seven vertices (0-simplices) and eight edges (1-simplices). We extend this complex by adding a 2-simplex, the triangle} $T=\{b,c,e\}$, as shown in Figure \ref{fig:enter-label}. {To analyze the topological structure of this complex, we define a filter function} $ g : V \rightarrow \mathbb{N} $ based on the degree of each vertex, such that $ g(v) = \deg(v) $,  where $ \deg(v) $ denotes the  number of edges incident to vertex $ v $.
The degree values for each vertex are:
\[
{g(f) = g(g)=1,\quad g(a) = g(e) = 2,\quad g(c)= g(d) = 3,\quad g(b)  = 4.}
\]
\end{example}

\vspace{-9pt}

We construct the \textit{lower-star filtration} of the graph step by step. At each filtration step $ t $, we include all simplices $ \sigma \subseteq V $ such that $ \max_{v \in \sigma} g(v) \leq t $.

{Step $t=1$:} We begin with an empty complex $ G_0 = \emptyset $. At a filtration value of 1, we add vertices $f$ and $g$ since $g(f)=g(g)=1$. No edges are included at this stage. Thus, the complex is  $G_{1}=G_{0}\cup \{\{f\},\{g\}\}$.

{Step $t=2$:} We include vertices $a$ and $e$, as their degree values are $g(a)=g(e)=2$. No edges are added because all neighboring vertices have higher degree values. The complex is now $G_{2}=G_{1}\cup \{\{a\},\{e\}\}$.

{Step $t=3$:} We add vertices $c$ and $d$, with degree values of 3. At this step, several edges are also included: $(c,e)$, $(c,d)$, $(a,d)$, and $(d,f)$, as the maximum degree of their endpoints is at most 3. The complex becomes $G_{3}=G_{2}\cup \{\{c\},\{d\},\{c,e\},\{c,d\},\{a,d\},\{d,f\}\}$.

\begin{figure}[H]
    \centering
    \includegraphics[width=1.0\linewidth]{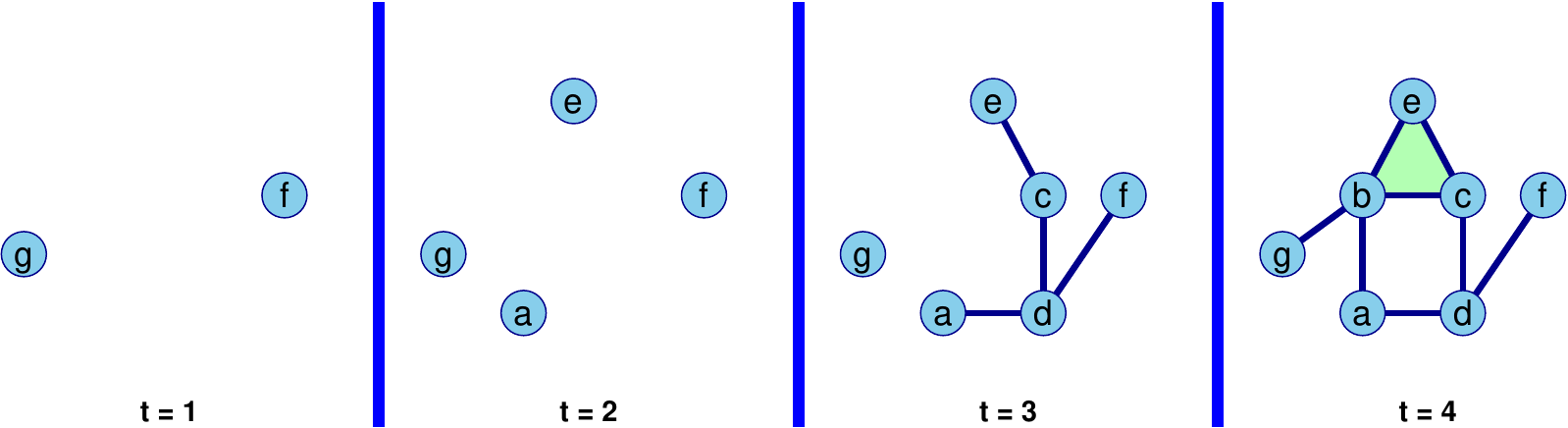}
    \caption{Illustration
 of the lower-star filtration at filtration steps $ t = 1, 2, 3, 4 $. Simplices are added according to the degree function $ g $ defined on the nodes of graph $ G $. Initially, only isolated vertices are added. As the filtration parameter $ t $ increases, edges and higher-order simplices (triangles) are included whenever all participating vertices have degrees less than or equal to $ t $.}
    \label{fig:lowerstar-final}
\end{figure}

{Step $t=4$:} Finally, we add vertex $b$ and all remaining simplices. All edges with a maximum vertex degree of 4-$(a,b)$, $(b,c)$, $(b,e)$, and $(b,g)$-are added. The 2-simplex, the triangle $\{b,c,e\}$, is also added since its maximum vertex degree is 4. The final complex is $G_{4}=G_{3}\cup \{\{b\},\{a,b\},\{b,c\},\{b,e\},\{b,g\},\{b,c,e\}\}$. Figure \ref{fig:lowerstar-final} visually demonstrates how the scalar function $ g $ induces a lower-star filtration on the graph $ G $, as  described above.

\begin{figure}[H]
    \centering

    \begin{minipage}[c]{0.48\textwidth}
        \centering
        \includegraphics[width=\linewidth]{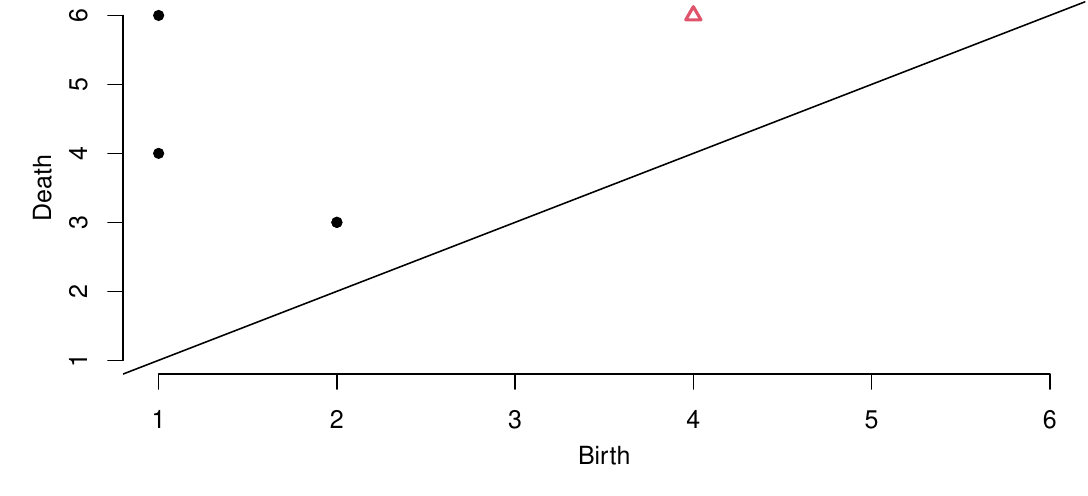}
        \caption{Visual representation of the persistence diagram
        shown in Table~\ref{tab:lowerstar_pd}, where each point
        corresponds to a topological feature characterized by
        its birth and death values.}
        \label{fig:placeholder}
    \end{minipage}
    \hfill
    \begin{minipage}[c]{0.48\textwidth}
        \centering
        \captionof{table}{Persistence diagram for the lower-star
        filtration of the graph in Figure~\ref{fig:enter-label}.
        Each row represents a topological feature with its
        dimension, birth time, and death time. Infinite death
        values indicate features that persist throughout
        the entire filtration.}
        \label{tab:lowerstar_pd}

        \renewcommand{\arraystretch}{1.2}
        \setlength{\tabcolsep}{12pt}

        \begin{tabular}{ccc}
            \toprule
            \textbf{Dimension} & \textbf{Birth} & \textbf{Death} \\
            \midrule
            $0$ & 1 & $\infty$ \\
            $0$ & 1 & 4 \\
            $0$ & 2 & 3 \\
            $0$ & 2 & 3 \\
            $1$ & 4 & $\infty$ \\
            \bottomrule
        \end{tabular}
    \end{minipage}

\end{figure}

{To compute the PD, we observe how topological features evolve as the filtration parameter} $ t $ increases. At $ t = 1 $, {two connected components are born; one persists indefinitely, while the other dies at }$ t = 3 $ {when it merges with another component through the formation of an edge (a 1-simplex). At} $ t = 2 $, two more connected components are born; both of these die at $ t = 3 $ as they merge with existing components via edge formation. At $ t = 4 $, {four more edges and a triangle (2-simplex) are added, which results in the birth of a loop. Since this loop is never filled in by a triangle at any later stage in the filtration, it persists indefinitely and is assigned an infinite death time in the PD. The PD resulting from the lower-star filtration is visualized in} Figure \ref{fig:placeholder} and detailed numerically in Table \ref{tab:lowerstar_pd}. {In practical applications, features with infinite death values-representing topological structures that never disappear-are handled differently depending on the problem at hand. They are often either excluded from the diagram or assigned a suitable finite constant to facilitate analysis and computation. However, in scenarios where the PD is used directly (e.g., with kernel methods or distance-based comparisons), it is often preferable to retain infinite death values, as they may carry meaningful topological information about essential or dominant features of the space.}

\subsection{Distance Metrics and Stability of Persistence Diagrams}

{A standard metric for comparing PDs is the $p$-Wasserstein distance} ($W_{p}$), defined for $p\ge 1$.{ Given two finite PDs, $D_{1}$ and $D_{2}$, the $p$-Wasserstein distance is} defined as:
\begin{equation}
W_{p}(D_{1},D_{2})=\left(\inf _{\gamma }\sum _{u\in D_{1}}\|u-\gamma (u)\|_{\infty }^{p}\right)^{1/p}
\end{equation}
{where $\|\cdot \|_{\infty }$ is the $L_{\infty }$ norm on $\mathbb{R}^{2}$, and $\gamma $ ranges over all bijections (matchings) between} {the diagrams. To ensure a perfect matching always exists, both $D_{1}$ and $D_{2}$ are augmented} {with countably infinite copies of the diagonal, $\Delta =\{(b,d)\in \mathbb{R}^{2}\mid b=d\}$. This allows}{ unmatched off-diagonal points to be paired with their orthogonal projections onto the diagonal, effectively absorbing topological noise. The bottleneck distance} ($d_{B}$) {is a special case of this metric. It measures the maximum distance between matched points under the best possible bijection:}
\begin{equation}
d_{B}(D_{1},D_{2})=\inf _{\gamma }\sup _{u\in D_{1}}\|u-\gamma (u)\|_{\infty }.
\end{equation}

The bottleneck distance is also the limit of the $p$-Wasserstein distance as $p\rightarrow \infty $. {The fundamental stability of persistence diagrams is guaranteed by these metrics. The stability theorem, first established for the bottleneck distance by Cohen-Steiner, Edelsbrunner, and Harer}~\cite{cohen2007stability}, {states that small perturbations in the scalar functions used to construct filtrations lead to only small changes in the resulting PDs. Formally, if} $f$ and $g$ are two tame functions (A function $ f :  \mathbb{X} \to \mathbb{R} $ is \emph{tame} if it has a finite 
number of homological critical values and the homology groups 
$ H_{k}(f^{-1}((-\infty, a])) $ are finite-dimensional for all 
$ k \in \mathbb{Z} $ and  $a \in \mathbb{R}$~\cite{cohen2007stability}), the bottleneck distance between their PDs is bounded by the sup-norm difference between the functions:
\begin{equation}
d_{B}(D(f),D(g))\le \|f-g\|_{\infty }.
\end{equation}
{This robustness holds for the $P$-Wasserstein} {distances as well, providing a principled way to quantify topological differences that is reliable even with noisy data.}

\subsection{Vector Summary of Persistence Diagrams}
\label{summary}

{Persistence diagrams offer a powerful way to summarize topological features in data. However, because the space of PDs is not a Hilbert space and PDs are multisets of varying sizes, they cannot be directly used in many statistical or machine learning methods. The metrics used to compare PDs, such as the Wasserstein or bottleneck distances, are also computationally expensive. Therefore, various vectorization techniques have been developed to transform PDs into finite-dimensional representations that are more compatible with standard algorithms.
These vectorization methods include persistence landscapes}~\cite{Bubenik_2020,berry2020functional,bubenik2015statistical,chazal2014stochastic}, persistence images~\cite{adams2016persistenceimagesstablevector}, persistence silhouettes~\cite{chazal2013stochasticconvergencepersistencelandscapes}, persistence entropy~\cite{atienza2020persistententropy}, persistence statistics~\cite{ali2023survey}, and Betti functions~\cite{chazal2021introduction, chung2022persistence, islambekov2024vector}. Many of these methods come with their own stability guarantees, ensuring that the vector representation remains robust to noise. In this work, we adopt three of these methods, providing a brief overview of their construction. Their implementations are available in the \emph{TDAvec} package~\cite{TDAvec2025}.

\begin{definition}[Persistence Statistics]
Given a PD, $ D = \{(b_i, d_i)\}_{i=1}^{N} $, where $ b_i $ and $ d_i $ denote the birth and death times of the $ i $th topological feature, \textit{persistence statistics} refer to a collection of descriptive measures derived from the birth values $ \{b_i\} $, death values $ \{d_i\} $, midpoint $ \left\{ \frac{b_i + d_i}{2} \right\} $, and lifespans (or persistences) $ \{d_i - b_i\} $. These statistics include the following:

\begin{itemize}
    \item Measures of central tendency and dispersion, such as the mean, standard deviation, median, interquartile range, range, and selected quantiles (10th, 25th, 75th, and 90th percentiles) for each of the four value types (birth, death, midpoint, and lifespan).
    \item The total number of points (or bars) in the diagram.
    \item {The entropy of the lifespan distribution is defined as} \begin{equation}
    H \;=\; -\sum_{i=1}^{N} \frac{l_i}{L}\,\log_{2}\!\left(\frac{l_i}{L}\right),
    \end{equation}
    where $ l_i = d_i - b_i $ represents the lifespan of the $ i $th feature and $ L = \sum_{i=1}^{N} l_i $ is the total sum\linebreak   of lifespans.
\end{itemize}
\end{definition}

\begin{definition}[Betti Function (also called a Betti curve)]
 Let $ D = \{(b_i, d_i)\}_{i=1}^N $  
be a PD associated with a fixed homological dimension $ k $, where each pair $ (b_i, d_i) $ represents the birth and death of a $ k $-dimensional topological feature. The \textbf{Betti function}
 $ \beta_k(t) $ is defined as:
\begin{equation}
\beta_k(t) = \sum_{i=1}^N \mathbf{1}_{[b_i, d_i)}(t),
\end{equation}
where $ \mathbf{1}_{[b_i, d_i)}(t) $ is the indicator function:
\begin{equation}
\mathbf{1}_{[b_i, d_i)}(t) = 
\begin{cases}
1 & \text{if } b_i \leq t < d_i, \\
0 & \text{otherwise}.
\end{cases}
\end{equation}

This function counts the number of $ k $-dimensional features that are alive at a given filtration value $ t $; specifically, the features that have emerged (been ``born'') but have not yet disappeared (``died'') at time $ t $. A common approach to vectorizing the Betti function is to evaluate it at each point in an increasing sequence of scale values $ \{t_1, t_2, \ldots, t_n\} $, resulting in a vector in $ \mathbb{R}^n $:
\[
(\beta_k(t_1),\, \beta_k(t_2),\, \ldots,\, \beta_k(t_n)) \in \mathbb{R}^n.
\]

\textls[-5]{Because the Betti function is piecewise constant and integrable, it can also be summarized by averaging over consecutive intervals. This produces a vector in $ \mathbb{R}^{n-1} $ of  the form:}
\begin{equation}
\left(
\frac{1}{\Delta t_1} \int_{t_1}^{t_2} \beta_k(t) \, dt,\,
\frac{1}{\Delta t_2} \int_{t_2}^{t_3} \beta_k(t) \, dt,\,
\ldots,\,
\frac{1}{\Delta t_{n-1}} \int_{t_{n-1}}^{t_n} \beta_k(t) \, dt
\right),
\end{equation}
where $ \Delta t_k = t_{k+1} - t_k $ for $ k = 1, \ldots, n-1 $. {While the two vectorization methods can yield similar results on a sufficiently dense grid, the interval-averaging method offers distinct advantages, particularly in terms of stability and robustness. The point-wise evaluation method (Equation (8)) is known to be unstable with respect to the 1-Wasserstein distance, as a small perturbation in the input data can lead to a large change in the output vector}~\cite{johnson2021instabilitybettisequencepersistent}. {In contrast, the interval-averaging vectorization considers the behavior of the Betti function over the entire intervals determined by neighboring scale values. This approach is more robust to small perturbations and captures the contribution of features that appear and disappear between sample points. This vectorization method can also be used to compute informative low-dimensional vectors based on a sparse grid of scale values, which helps mitigate noise and reduces the dimensionality of the feature representation}~\cite{islambekov2023fasttopologicalapproachpredicting}. {(See the experiment of Section 3.1 in~\cite{islambekov2023fasttopologicalapproachpredicting} which compares the two vectorization methods of the Betti function when the extracted
vector summaries are low-dimensional).}

\end{definition}

\begin{definition}[Generalized Persistence Landscape (generalized persistence landscape is a kernel-based variant of the standard persistence landscape; see~\cite{TDAvec2025} for more details and implementations)]
Let $D = \{(b_i, d_i)\}_{i=1}^N$ 
be a PD in a fixed homological dimension $ k $, where each pair  
$(b_i, d_i) \in \mathbb{R}^2$ represents the birth and death of a topological feature. The {$j$}th-order persistence landscape function is defined as follows:
\begin{equation*}
\lambda_j(t) = {j}\text {-}\max_{1 \leq i \leq N} \Lambda_i(t), \quad {j} = 1, \dots,k,
\end{equation*}where ${j} \max$ denotes the {$j$}th largest value among the set $\{\Lambda_1(t), \Lambda_2(t), \ldots, \Lambda_N(t)\}$. Each $\Lambda_i(t)$ is a piecewise-linear, triangular ``tent'' function defined as:
\begin{equation}
\Lambda_i(t) =
\begin{cases}
t - b_i & \text{if } t \in [b_i, \frac{b_i + d_i}{2}], \\
d_i - t & \text{if } t \in (\frac{b_i + d_i}{2}, d_i], \\
0 & \text{otherwise}.
\end{cases}
\end{equation}
Each triangle peaks at $ t = \frac{b_i + d_i}{2} $ with height $ \frac{d_i - b_i}{2} $, capturing the lifespan of the corresponding topological feature.

To obtain a finite-dimensional representation suitable for machine learning, each landscape function $ \lambda_k(t) $ is sampled at a predefined sequence of increasing filtration values $ \{ t_1, t_2, \ldots, t_n \} $, resulting in a vector of the form:

\[
\left( \lambda_k(t_1), \lambda_k(t_2), \ldots, \lambda_k(t_n) \right) \in \mathbb{R}^n.
\]

The \textbf{generalized persistence landscape} extends the traditional definition by replacing the sharp triangular functions $ \Lambda_i(t) $ with smooth kernel-based bump functions. Let $ K : \mathbb{R} \rightarrow \mathbb{R} $ be a symmetric kernel supported on the interval $ [-1, 1] $, and define its scaled version with bandwidth $ h > 0 $ as: $K_h(t) = \frac{1}{h} K\left( \frac{t}{h} \right).$

Then, for each point $ (b_i, d_i) $, the smoothed function $ \Lambda_i(t) $ is defined by:

\begin{equation}    
\Lambda_i(t) = 
\begin{cases}
\displaystyle \frac{d_i - b_i}{2K_h(0)} \cdot K_h\left(t - \frac{b_i + d_i}{2}\right) & \text{if } \left| \frac{t - \frac{b_i + d_i}{2}}{h} \right| \leq 1, \\
0 & \text{otherwise}.
\end{cases}
\end{equation}

As before, generalized landscape functions $ \lambda_k(t) $ are defined in the same way. Common choices for the kernel $ K $ include the triangle kernel, the Epanechnikov kernel, and the tricubic kernel, each offering different smoothness properties and bandwidth sensitivities. 
  
\end{definition}

\begin{figure}[ht]
    \centering
    \subfloat[\centering Persistence diagram]{%
        \includegraphics[width=0.48\linewidth]{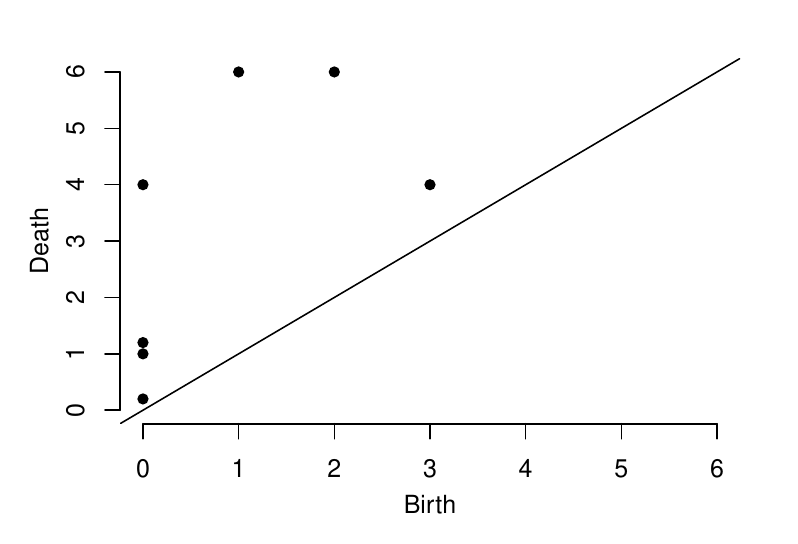}%
        \label{subfig:a}%
    }\hfill
    \subfloat[\centering Persistence Statistics]{%
        \includegraphics[width=0.48\linewidth]{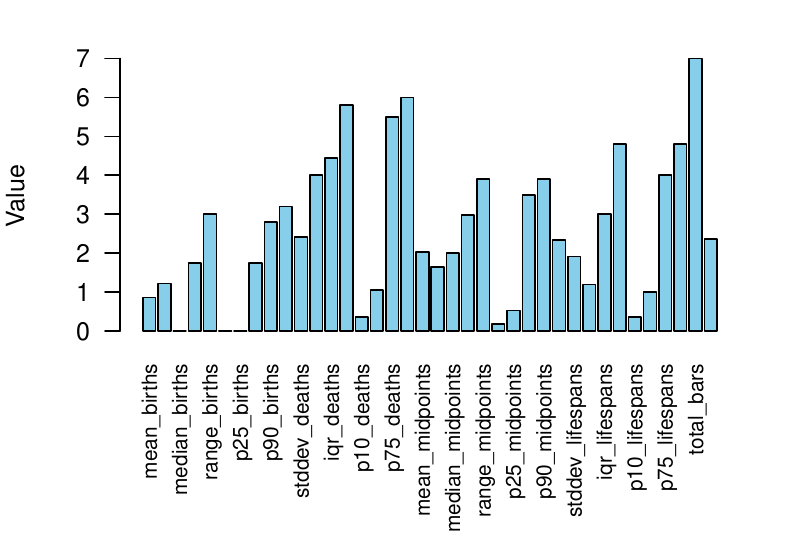}%
        \label{subfig:b}%
    }\\[1ex]
    \subfloat[\centering Betti Function]{%
        \includegraphics[width=0.48\linewidth]{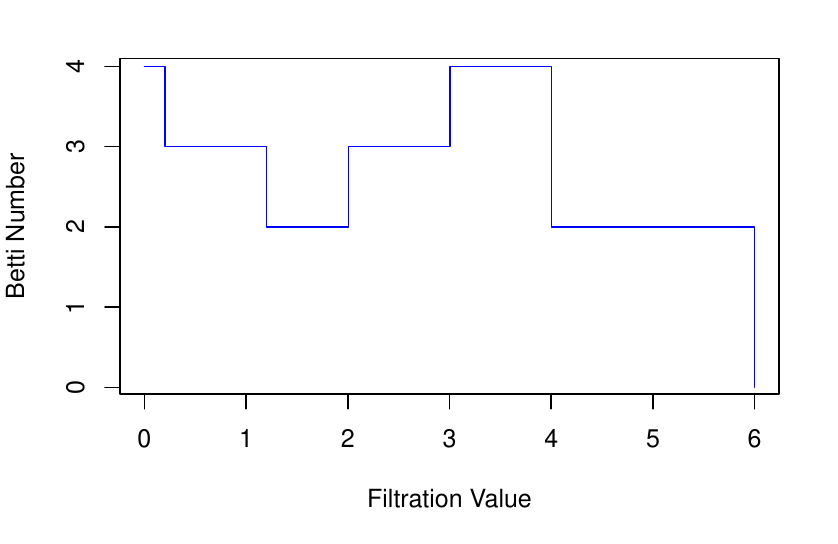}%
        \label{subfig:c}%
    }\hfill
    \subfloat[\centering Generalized Persistence Landscape]{%
        \includegraphics[width=0.48\linewidth]{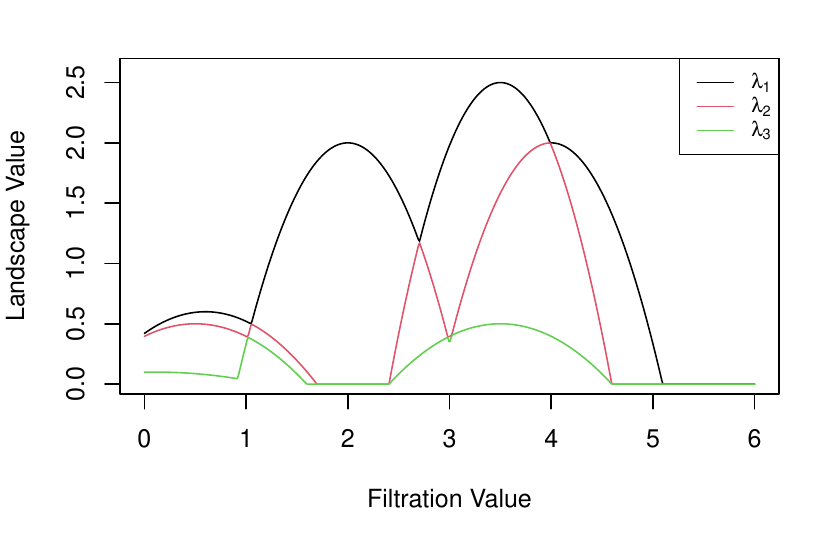}%
        \label{subfig:d}%
    }
      \caption{(\textbf{a}) illustrates a sample PD, while (\textbf{b}--\textbf{d}) display its corresponding vectorized summaries: persistence statistics, Betti function, and generalized persistence landscape, respectively. The generalized persistence landscape in (\textbf{d}) was constructed using the Epanechnikov kernel.}
    \label{fig:fig}
\end{figure}

Figure \ref{fig:fig} {provides} a graphical illustration of these summary functions:
(a) \textit{Persistence diagram}
—displays the birth and death times of topological features as points in the plane.
(b) \textit{Persistence statistics}—summarizes information from the persistence diagram using numerical descriptors such as mean, standard deviation, and maximum lifetime.
(c) \textit{Betti function}—shows how the number of topological features (e.g., connected components or loops) changes over the filtration process.
(d) \textit{Generalized persistence landscape}—{encodes topological features into functional representations that are suitable for statistical comparison and integration into machine learning models. Unlike classical persistence landscapes, the generalized version may include smoothing or nonlinear transformations and is not necessarily piecewise-linear.}

\vspace{-9pt}
\section{Proposed Nested Ensemble Learning Framework} 
\label{sec:methodology}

In this section, we present our proposed \emph{nested ensemble learning framework}, {designed to integrate topological features extracted from graph data using a hierarchical stacking approach. Stacking is an ensemble technique in which the predictions of multiple base models are used as input features for one or more higher-level meta-learners, which aim to learn optimal combinations of these predictions. Our framework extends traditional stacking by introducing additional layers, resulting in a three-level architecture: Level-0 (Base Models), Level-1 (Filter-Specific Ensembles), and Level-2 (Cross-Filter Ensemble). Level-1 meta-learners combine the outputs of base models within each filter to form filter-specific ensembles. The final nested ensemble prediction is then produced by a Level-2 meta-learner, which learns to optimally weight and integrate predictions from the different filter-specific ensembles. The specific choice of meta-learners at both levels depends on the complexity and characteristics of the data.}

Let $G=(V,E)$ denote a graph with node set $V$ and edge set $E$. For each node $v\in V$, we compute three distinct node attributes (filter functions): Closeness centrality ($f_{\text{closeness}}(v)$), Degree-2 centrality ($f_{\text{degree2}}(v)$), and Betweenness centrality ($f_{\text{betweenness}}(v)$).

For each graph $ G_i $ and each filter function $ f \in \{ f_{\text{closeness}}, f_{\text{degree2}}, f_{\text{betweenness}} \} $, we compute a persistence diagram $ D_i^{(f)} $ via a lower-star filtration. This yields:
\begin{equation}    
D^{(f)} = \{ D_1^{(f)}, D_2^{(f)}, \ldots, D_N^{(f)} \},
\end{equation} capturing the topological features of all graphs under filter $ f $.

Each $ D_i^{(f)} $ is vectorized into a fixed-dimensional feature representation using summary functions (e.g., persistence statistics, Betti curves, generalized persistence landscapes). This produces a feature matrix:
\begin{equation}
X^{(f)} \in \mathbb{R}^{N \times d_f},
\end{equation}
where $ N $ is the number of samples (graphs) and $ d_f $ is the feature dimension associated with filter $ f $. This is performed separately for the training and test sets, yielding
\[
X_{\text{train}}^{(f)}, X_{\text{test}}^{(f)} \quad \text{along side labels} \quad y_{\text{train}}, y_{\text{test}}.
\]

We propose a three-level ensemble learning procedure, described as follows:

\begin{enumerate}

\item \textbf{Base Model Training (Level-0):} 

For each filter $f\in \{f_{\text{closeness}},f_{\text{degree2}},f_{\text{betweenness}}\}$, we independently train $M$ diverse base models $\{h_{1}^{(f)},h_{2}^{(f)},\dots ,h_{M}^{(f)}\}$ on the training feature matrix $X_{\text{train}}^{(f)}$. Each base model, $h_{m}^{(f)}$, is a learning algorithm (e.g., a neural network, support vector machine, or gradient boosting model) that maps $\mathbb{R}^{d_{f}}\rightarrow \mathbb{R}$. The predictions from each base model are then given by:\begin{equation}
\hat{y}_{j, \text{train}}^{(f)} 
= h_j^{(f)}\!\bigl( X_{\text{train}}^{(f)} \bigr),
\quad j = 1, \dots, M.
\end{equation}

\item \textbf{Filter-specific Ensemble (Level-1):}
For each filter function $ f $, the predictions from the $ M $ base models are aggregated into a matrix:
\begin{equation}
Z_{\text{train}}^{(f)} = 
\begin{bmatrix}
\hat{y}_{1,\text{train}}^{(f)} &
\hat{y}_{2,\text{train}}^{(f)} &
\cdots &
\hat{y}_{M,\text{train}}^{(f)}
\end{bmatrix}
\in \mathbb{R}^{N \times M}.
\end{equation}

{A meta-learner $ g^{(f)} : \mathbb{R}^{M} \rightarrow \mathbb{R} $} {is then trained to combine these base-model predictions. This meta-learner takes the outputs of the Level-0 models as input features and learns the optimal way to assign weights to each base model based on their predictions, producing a single, refined prediction for each sample. The resulting predictions for the training data are:}
\begin{equation}
\tilde{y}_i^{(f)} = g^{(f)}\bigl(Z_i^{(f)}\bigr), \quad i = 1, \dots, N.
\end{equation}

The choice of meta-learner $ g^{(f)} $ depends on the complexity of the task; simple models like logistic regression may suffice, while more flexible methods such as tree-based models or shallow neural networks can be used when needed.

These refined outputs form the filter-specific ensemble prediction:
\begin{equation}
\tilde{y}_{\text{train}}^{(f)} = 
\left[ \tilde{y}_1^{(f)}, \tilde{y}_2^{(f)}, \dots, \tilde{y}_N^{(f)} \right]^\top.
\end{equation}

This process is repeated for each filter function, resulting in three separate Filter-specific Ensembles:
\[
E^{(\text{closeness})}, \quad
E^{(\text{degree2})}, \quad
E^{(\text{betweenness})}.
\]

\item \textbf{Cross-filter Ensemble (Level-2):}
At the second level, we combine the outputs from the three filter-specific ensembles into a new training set. For the training data, we form a matrix:
\begin{equation}
W_{\text{train}} = 
\begin{bmatrix}
\tilde{y}_1^{(\text{closeness})} & \tilde{y}_1^{(\text{degree2})} & \tilde{y}_1^{(\text{betweenness})} \\
\vdots & \vdots & \vdots \\
\tilde{y}_N^{(\text{closeness})} & \tilde{y}_N^{(\text{degree2})} & \tilde{y}_N^{(\text{betweenness})}
\end{bmatrix}
\in \mathbb{R}^{N \times 3}
\end{equation}

Similarly, for the test data we construct:
\begin{equation}
W_{\text{test}} = 
\begin{bmatrix}
\tilde{y}_{N+1}^{(\text{closeness})} & \tilde{y}_{N+1}^{(\text{degree2})} & \tilde{y}_{N+1}^{(\text{betweenness})} \\
\vdots & \vdots & \vdots \\
\tilde{y}_{N+T}^{(\text{closeness})} & \tilde{y}_{N+T}^{(\text{degree2})} & \tilde{y}_{N+T}^{(\text{betweenness})}
\end{bmatrix}
\in \mathbb{R}^{T \times 3}
\end{equation}

Another meta-learner $\mathcal{G}:\mathbb{R}^{3}\rightarrow \mathbb{R}$ is trained on the outputs of the filter-specific ensemble, using the training set $(W_{\text{train}},y_{\text{train}})$. As a distinct learning algorithm, $\mathcal{G}$ takes as input a three-dimensional vector containing the predictions from the filter-specific ensemble ($E^{(\text{closeness})}$, $E^{(\text{degree2})}$, $E^{(\text{betweenness})}$) and learns to assign adaptive weights to these inputs. The final nested ensemble prediction is then produced:

\begin{adjustwidth}{-\extralength}{0cm}
\begin{equation}\hat{y}{\text{final}} = \mathcal{G}\left( \left[ E{\text{closeness}}(X^{(\text{closeness})}), E_{\text{degree2}}(X^{(\text{degree2})}), E_{\text{betweenness}}(X^{(\text{betweenness})}) \right] \right).\end{equation}
\end{adjustwidth}
\vspace{6pt}

The optimization objective is to minimize the generalization error:
\begin{equation}
\min_{\mathcal{G}, \{ g^{(f)} \}, \{ h_j^{(f)} \}} \mathbb{E}_{(X,y)} \left[ \mathcal{L}\left( y, \hat{y}_{\text{final}} \right) \right],
\end{equation}
where $ \mathcal{L} $ is a suitable loss function (e.g., mean squared error for regression, cross-entropy for classification).

\end{enumerate}

\begin{figure}[H]
    \centering
    \includegraphics[width=1.0\textwidth, height=2.9in]{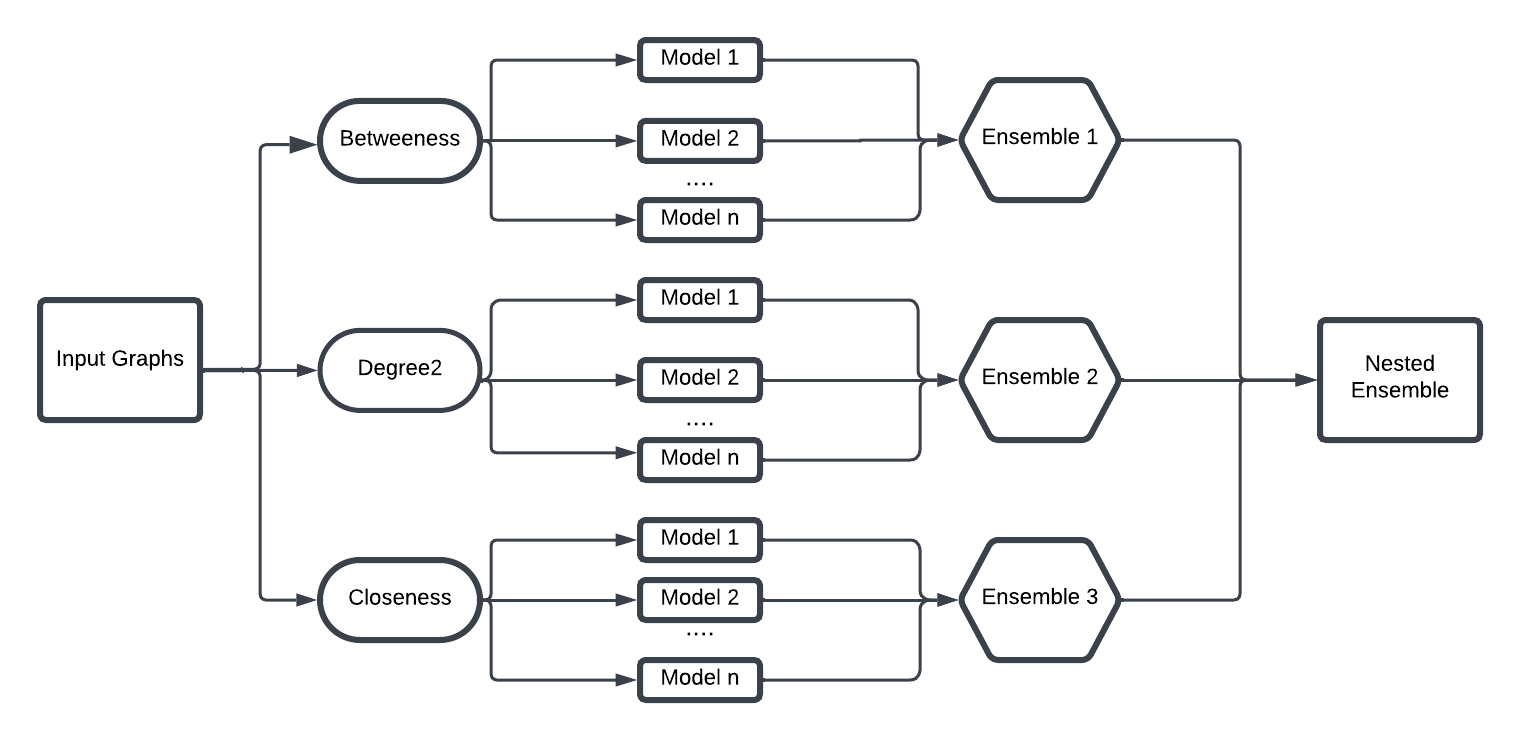}
    \caption{This figure illustrates the nested ensemble architecture. The process begins with graphs as input features, from which persistence diagrams are computed using lower-star filtrations induced by filter functions. The filter functions are node attributes: closeness, betweenness, and a custom variant of degree centrality called degree2. For each filter function, $M$ base models are trained independently. Their outputs are then combined by Level-1 meta-learners into filter-specific ensembles. Finally, a Level-2 meta-learner integrates these ensembles to produce the final nested ensemble prediction.}
    \label{fig:nested_ensemble}
\end{figure}

The optimization process trains the Level-2 meta-learner to assign dynamic, context-aware weights to each filter-specific ensemble prediction. In this manner, the framework effectively synthesizes complementary topological signals into a single, unified output while minimizing prediction error. This integration strategy is designed to preserve the local discriminative power of individual filter ensembles while simultaneously capturing the global relationships between different topological perspectives. As a result, the approach yields more robust predictions than any single-filter method. A schematic overview of the full nested ensemble architecture is provided in Figure~\ref{fig:nested_ensemble}.

 \vspace{-9pt}


\section{Experimental Setup}
\label{sec:results}
 
In this section, we evaluate the performance of the proposed nested ensemble framework by comparing it against both the individual base models and the filter-specific ensembles (i.e., stacking ensembles).
 The evaluation is carried out on both simulated and real-world datasets, across classification and regression tasks. For the classification experiments, we further benchmark our approach against several state-of-the-art methods to highlight its competitive effectiveness. To vectorize the PDs, we use three summary functions described in Section \ref{summary}: persistence statistics (PS), the Betti function (BF), and generalized persistence landscapes (PLs). {For both the BF and PL representations, we compute values over a fixed scale sequence of 100 evenly spaced points ranging from 0 to 2. In the case of generalized persistence landscapes, we additionally apply a kernel smoothing procedure based on an Epanechnikov kernel with a bandwidth parameter set to} $h=0.2$.

In the simulation study, we generate synthetic graphs using three well-established models: the Erdős–Rényi model (ERM)~\cite{erdos1959random}, the stochastic block model (SBM)~\cite{Holland1983} and the Watts–Strogatz model~\cite{watts1998smallworld}. Graphs generated from the Erdős–Rényi model and stochastic block model are employed to evaluate classification performance, while those generated by the Watts–Strogatz model are used to assess regression performance.
 
For the regression tasks, we compare the performance of the proposed nested ensemble model against both individual base models and stacking ensembles, using two distinct feature extraction approaches. The first approach leverages topological features derived from persistent homology, where lower-star filtrations are constructed by using node attributes as filter functions. The second approach is purely non-topological, relying solely on quantile-based summaries of the node attributes. To further validate our methodology, we evaluate our model on four widely used benchmark datasets for graph classification—\emph{IMDB-BINARY}, \emph{IMDB-MULTI}, \emph{REDDIT-BINARY}, and \emph{PROTEINS}—all sourced from the TUDataset collection~\cite{kersting2016benchmark}. These datasets, which are constructed from social and biological network structures, are frequently used in fields such as biomedical and biological network analysis. Descriptive statistics and additional details for each dataset are provided in Table \ref{tab:real_data_summary}. We also compare the performance of our proposed approach against several established graph classification methods~\cite{damke2020novelhigherorderweisfeilerlehmangraph,errica2022faircomparisongraphneural,ivanov2018anonymous,xu2019powerfulgraphneuralnetworks,zhang2018end}. {All analyses were conducted} {using the R statistical software environment, and all machine learning models were trained using default hyperparameter settings unless otherwise specified.}

\subsection{Simulation Studies}
 
To evaluate the performance and robustness of our proposed nested ensemble framework across diverse network structures, we conducted two sets of simulation experiments. The first experiment uses Erdős–Rényi graphs to model homogeneous random networks, while the second employs stochastic block models (SBMs) to generate networks with well-defined community structures. Together, these experiments highlight the versatility of our approach across a broad spectrum of graph topologies.

\subsubsection{Experiment on Erdős--Rényi Graphs}
 
{Before evaluating the performance of the proposed model on Erdős--Rényi graphs, we first assess the computational cost and scalability of our method for extracting topological features from graph data. This evaluation is carried out under two controlled conditions. The first condition varies the number of graphs} {($N$) while keeping the edge probability ($p$) for} each class fixed. The second condition fixes the number of graphs ($N$) and varies the edge probabilities ($p$) {across classes. For each generated graph, node-level attributes—including degree2, closeness, and betweenness—are first computed to serve as filter functions. Subsequently, lower-star filtrations are constructed based on these attributes, and PDs are computed from the filtrations. In all cases, the number of nodes in each graph is uniformly sampled from the interval} $[80, 120]$. The results are presented in Table~\ref{tab:runtime_random}a,b, which provide the median run-time costs (measured in seconds and computed based on ten repeated simulations) for these two studies. Table~\ref{tab:runtime_random}a demonstrates that the median run-time for computing PDs scales efficiently and nearly linearly with the number of graphs $N$ when graph density is fixed.

\begin{table}[H]
\centering
\caption{Comparison of median runtime (in seconds) for computing
persistence diagrams (PDs) from Erdős--Rényi graphs under two
experimental setups: varying the number of graphs with fixed edge
probabilities (left), and fixing the number of graphs while randomly
sampling edge probabilities (right). Each run includes computing
node attributes (closeness, betweenness, and degree2), constructing
lower-star filtrations, and extracting PDs.}
\label{tab:runtime_random}

\begin{tabularx}{\textwidth}{@{}*{8}{>{\centering\arraybackslash}X}@{}}
\toprule
\multicolumn{4}{c}{\textbf{(a) Varying $N$, Fixed Probabilities}} &
\multicolumn{4}{c}{\textbf{(b) Fixed $N=100$, Random Probabilities}} \\
\midrule

$\boldsymbol{N}$ &
$\boldsymbol{p_1}$ &
$\boldsymbol{p_2}$ &
\textbf{Runtime (s)} &
\textbf{Trial} &
$\boldsymbol{p_1}$ &
$\boldsymbol{p_2}$ &
\textbf{Runtime (s)} \\
\midrule

100  & 0.05 & 0.075 & 1.70  & 1  & 0.2326 & 0.1116 & 22.30  \\
200  & 0.05 & 0.075 & 3.02  & 2  & 0.4987 & 0.2280 & 202.84 \\
300  & 0.05 & 0.075 & 5.26  & 3  & 0.0601 & 0.0206 & 2.39   \\
400  & 0.05 & 0.075 & 5.97  & 4  & 0.3972 & 0.4311 & 243.71 \\
500  & 0.05 & 0.075 & 7.71  & 5  & 0.0876 & 0.3472 & 86.04  \\
600  & 0.05 & 0.075 & 9.58  & 6  & 0.3606 & 0.2420 & 109.88 \\
700  & 0.05 & 0.075 & 10.33 & 7  & 0.2169 & 0.3658 & 103.83 \\
800  & 0.05 & 0.075 & 12.58 & 8  & 0.3118 & 0.0688 & 54.86  \\
900  & 0.05 & 0.075 & 14.29 & 9  & 0.1617 & 0.3902 & 97.27  \\
1000 & 0.05 & 0.075 & 15.88 & 10 & 0.4599 & 0.3481 & 221.22 \\

\bottomrule
\end{tabularx}
\end{table}

{For example, processing 100 graphs took approximately 1.70 s, while 1000 graphs required about 15.88 s. In contrast, }Table~\ref{tab:runtime_random}b {reveals that run-time varies considerably with graph complexity. With the number of graphs fixed at 100, randomly sampling the edge probabilities ($p_1$, $p_2$) resulted in median run-times ranging from 2.39 to 221.22 s. This demonstrates the pipeline's computational efficiency with respect to dataset size, as well as its sensitivity to graph structure and edge density. We now proceed to evaluate its performance in a classification context.}

 In the first experiment, we generate synthetic graphs using the Erdős--Rényi model. The number of nodes $n$ in each graph is uniformly sampled from the interval $[80, 120]$. Each graph is assigned to one of two classes based on its edge probability; Class~1 uses $p_1 = 0.35$, while Class~2 uses $p_2 = 0.30$. The resulting dataset consists of 300 training graphs (150 per class) and 100 test graphs (50 per class). A summary of the average network properties, including density, average degree (AD), average shortest path distance (ASD), diameter, and transitivity, is provided in Table~\ref{tab:er_network_properties}. These metrics underscore the structural differences between the two classes.

\begin{table}[H]
\centering
\caption{Mean network properties for ERM graphs with
different edge probabilities. AD denotes average degree,
and ASD denotes average shortest-path distance.}
\label{tab:er_network_properties}

\begin{tabularx}{\textwidth}{@{}Xccccc@{}}
\toprule
\textbf{Class} &
\textbf{Density} &
\textbf{AD} &
\textbf{ASD} &
\textbf{Diameter} &
\textbf{Transitivity} \\
\midrule

Class 1 & 0.35 & 34.72 & 1.65 & 2.02 & 0.35 \\
Class 2 & 0.30 & 29.79 & 1.70 & 2.35 & 0.30 \\

\bottomrule
\end{tabularx}
\end{table}

We compute PDs using the lower-star filtration induced by a function $ g $, whose values are normalized to the interval $[0, 1]$. As filter functions for $ g $, we use closeness, betweenness, {and the custom-defined variant of degree centrality defined as degree 2 in Section
 \ref{section2.1.1}}. PDs are computed for each of these filter functions on both the training and test sets. These features, which capture the birth and death of topological features throughout the filtration process, are then vectorized. For each filter function, we train three base machine learning models: Bagging, Random Forest, and Boosting. These base models are then combined using a {generalized linear model (GLM) as a Level-1 meta-learner, which optimally weights their predictions to produce a single ensemble model for each filter function. Finally, a Random Forest (RF) is used as a Level-2 meta-learner }to integrate the outputs of the three filter-specific ensembles, constructing the final nested ensemble model.
 Table~\ref{tab:tda_results} reports the classification accuracies (mean~$\pm$~standard deviation) obtained from the simulated study.

 \begin{table}[H]
\centering
\caption{Classification accuracy (\%, mean $\pm$ standard deviation
from 10-fold cross-validation) across homology dimensions
($H_0$, $H_1$) using three summary functions (PS, BF, PL).
Features are extracted from three filter functions: degree2,
betweenness, and closeness. For each filter function, Bagging,
Random Forest, and Boosting models are trained and combined
into a stacked ensemble. These stacked ensembles are then
integrated into a final nested ensemble model.}
\label{tab:tda_results}

\begin{tabularx}{\textwidth}{@{}lXcccc@{}}
\toprule
\textbf{Summary} &
\textbf{Filter} &
\textbf{Bagging} &
\textbf{RF} &
\textbf{Boosting} &
\textbf{Stacked} \\
\midrule

\multirow[c]{4}{*}{\textbf{PS}}
& Closeness $H_0$
& $59.3 \pm 4.8$
& $60.7 \pm 4.1$
& $61.6 \pm 6.7$
& $61.6 \pm 5.9$ \\

& Betweenness $H_0$
& $64.6 \pm 3.0$
& $66.8 \pm 2.7$
& $67.2 \pm 3.5$
& $67.3 \pm 4.3$ \\

& Degree 2 $H_0$
& $72.3 \pm 2.8$
& $72.5 \pm 3.0$
& $72.0 \pm 3.3$
& $72.8 \pm 4.1$ \\

& Nested Ensemble $H_0$
& -- & -- & --
& $\boldsymbol{74.6 \pm 5.9}$ \\

\midrule

\multirow[c]{4}{*}{\textbf{BF}}
& Closeness $H_0$
& $51.5 \pm 3.7$
& $53.7 \pm 2.9$
& $54.4 \pm 5.0$
& $53.0 \pm 3.5$ \\

& Betweenness $H_0$
& $66.6 \pm 4.8$
& $66.9 \pm 6.2$
& $68.3 \pm 6.0$
& $68.8 \pm 5.5$ \\

& Degree 2 $H_0$
& $70.1 \pm 6.2$
& $70.5 \pm 5.7$
& $69.9 \pm 6.5$
& $69.5 \pm 6.2$ \\

& Nested Ensemble $H_0$
& -- & -- & --
& $\boldsymbol{72.1 \pm 5.7}$ \\

\midrule

\multirow[c]{4}{*}{\textbf{PL}}
& Closeness $H_0$
& $55.9 \pm 5.3$
& $58.3 \pm 5.3$
& $58.3 \pm 2.8$
& $56.6 \pm 4.3$ \\

& Betweenness $H_0$
& $65.2 \pm 4.2$
& $65.4 \pm 5.7$
& $65.5 \pm 5.2$
& $66.5 \pm 5.3$ \\

& Degree 2 $H_0$
& $70.9 \pm 6.4$
& $73.5 \pm 6.2$
& $72.9 \pm 6.6$
& $73.5 \pm 7.0$ \\

& Nested Ensemble $H_0$
& -- & -- & --
& $\boldsymbol{75.8 \pm 8.6}$ \\

\midrule

\multirow[c]{4}{*}{\textbf{PS}}
& Closeness $H_1$
& $90.8 \pm 3.6$
& $91.0 \pm 3.9$
& $91.8 \pm 2.6$
& $92.0 \pm 2.4$ \\

& Betweenness $H_1$
& $93.8 \pm 3.2$
& $93.8 \pm 3.3$
& $94.1 \pm 3.2$
& $94.1 \pm 2.6$ \\

& Degree 2 $H_1$
& $93.3 \pm 2.4$
& $93.3 \pm 2.6$
& $94.0 \pm 2.3$
& $93.8 \pm 2.4$ \\

& Nested Ensemble $H_1$
& -- & -- & --
& $\boldsymbol{96.5 \pm 3.2}$ \\

\midrule

\multirow[c]{4}{*}{\textbf{BF}}
& Closeness $H_1$
& $82.7 \pm 3.2$
& $82.8 \pm 3.8$
& $69.1 \pm 3.7$
& $83.3 \pm 3.1$ \\

& Betweenness $H_1$
& $92.9 \pm 2.6$
& $94.0 \pm 2.5$
& $93.5 \pm 2.3$
& $94.2 \pm 2.3$ \\

& Degree 2 $H_1$
& $87.2 \pm 4.2$
& $88.3 \pm 3.8$
& $88.2 \pm 3.9$
& $89.1 \pm 4.2$ \\

& Nested Ensemble $H_1$
& -- & -- & --
& $\boldsymbol{95.1 \pm 2.6}$ \\

\midrule

\multirow[c]{4}{*}{\textbf{PL}}
& Closeness $H_1$
& $72.3 \pm 4.9$
& $72.5 \pm 3.7$
& $73.9 \pm 2.9$
& $75.1 \pm 3.3$ \\

& Betweenness $H_1$
& $92.8 \pm 2.1$
& $92.5 \pm 2.5$
& $93.7 \pm 2.9$
& $93.0 \pm 2.8$ \\

& Degree 2 $H_1$
& $80.4 \pm 0.4$
& $81.1 \pm 4.2$
& $81.4 \pm 3.4$
& $81.1 \pm 3.8$ \\

& Nested Ensemble $H_1$
& -- & -- & --
& $\boldsymbol{94.7 \pm 3.0}$ \\

\bottomrule
\end{tabularx}
\end{table}

 The results clearly demonstrate that the proposed nested ensemble model consistently outperforms both individual base models and stacked ensembles, regardless of the summary
 function (PS, BF, PL) or homology dimension ($H_0$ or $H_1$) considered. For example, in the $H_0$ dimension with Persistence Statistics (PS), the nested ensemble achieved an accuracy of 74.6, representing a $2.1\%$ improvement over the best-performing base model (72.5) and a $1.8\%$ gain over the top-stacked ensembles (72.8). The most substantial improvement in $H_0$ was observed with the Persistence Landscape (PL), where the nested ensemble reached an accuracy of 75.8, outperforming the best individual base model (72.9) by $2.9\%$ and the best stacked ensembles (73.5) by $2.3\%$. Performance gains were even more pronounced in the $H_1$ dimension. Using PS, the nested ensemble achieved an accuracy of 96.5, exceeding both the best individual base model and the stacked ensembles (each at 94.1) by $2.4\%$. With the Betti functions (BFs), the nested ensemble attained 95.1, surpassing the best stacked ensembles (94.2) by $0.9\%$ and the top-performing individual base model (94.6) by $0.5\%$. Finally, with PL, the nested ensemble achieved 94.7, yielding improvements of $1.0\%$ and $0.9\%$ over the best base model (93.7) and stacked ensembles (93.8), respectively.

\subsubsection{Experiment on Stochastic Block Model Graphs}
 
In the second experiment, we explored networks with inherent community structure, modeled using the stochastic block model (SBM). In this framework, a community is conceptualized as a subset of nodes that are densely connected internally, with relatively sparse connections between different communities.

To generate these networks, we use Erdős--Rényi models $ G(n, p_{\text{high}}) $ and $ G(n, p_{\text{low}}) $ to represent within- and between-community connections, respectively. We considered three different SBM configurations, with the number of communities set to $ k_0 = 2, 5, $ and $ 10 $, each of roughly equal size. For each configuration, we generate 150 networks with edge probabilities $(p_{\text{high}}, p_{\text{low}}) = (0.8, 0.1)$, where the number of nodes $ n $ is uniformly sampled from $[80, 120]$. Table \ref{tab:sbm_network_properties} summarizes the mean network properties for these SBM-generated graphs, demonstrating the increasing heterogeneity in network characteristics as the number of communities increases. These datasets are then split into training and testing sets using a 75--25\% ratio.

\begin{table}[H]
\centering
\caption{Mean network properties of SBM graphs for
$k_0 \in \{2,5,10\}$. AD denotes average degree,
and ASD denotes average shortest-path distance.}
\label{tab:sbm_network_properties}

\begin{tabularx}{\textwidth}{@{}Xccccc@{}}
\toprule
$\boldsymbol{k_0}$ &
\textbf{Density} &
\textbf{AD} &
\textbf{ASD} &
\textbf{Diameter} &
\textbf{Transitivity} \\
\midrule

2  & 0.45 & 44.17 & 1.55 & 2.41 & 0.66 \\
5  & 0.23 & 22.92 & 1.79 & 3.00 & 0.41 \\
10 & 0.17 & 16.50 & 1.92 & 3.00 & 0.26 \\

\bottomrule
\end{tabularx}
\end{table}

We computed PDs for both the training and testing data for each filter function using a lower-star filtration, and then vectorized these diagrams with three topological summary functions (PS, BF, PL). The results are shown in Table \ref{tab:tda_results1}.

\begin{table}[H]
\centering
\caption{Classification accuracy (\%, mean $\pm$ standard deviation
from 10-fold cross-validation) across homology dimensions
($H_0$, $H_1$) using three topological summary functions
(PS, BF, PL). Features are extracted from three filter functions.
For each filter, models are trained and combined into a stacked
ensemble, which is then integrated into a final nested ensemble
using a meta-model.}
\label{tab:tda_results1}

\begin{tabularx}{\textwidth}{@{}lXcccc@{}}
\toprule
\textbf{Summary} &
\textbf{Filter} &
\textbf{Bagging} &
\textbf{RF} &
\textbf{Boosting} &
\textbf{Stacked} \\
\midrule

\multirow[c]{4}{*}{\textbf{PS}}
& Closeness $H_0$
& $89.2 \pm 3.0$
& $91.5 \pm 2.4$
& $90.8 \pm 2.5$
& $91.7 \pm 1.6$ \\

& Betweenness $H_0$
& $96.1 \pm 1.6$
& $96.3 \pm 1.4$
& $95.9 \pm 2.0$
& $95.9 \pm 1.8$ \\

& Degree 2 $H_0$
& $96.9 \pm 1.0$
& $97.1 \pm 1.6$
& $97.3 \pm 1.3$
& $97.2 \pm 1.4$ \\

& Nested Ensemble $H_0$
& -- & -- & --
& $\boldsymbol{99.7 \pm 0.4}$ \\

\midrule

\multirow[c]{4}{*}{\textbf{BF}}
& Closeness $H_0$
& $84.3 \pm 2.4$
& $84.1 \pm 2.7$
& $84.9 \pm 2.8$
& $85.6 \pm 2.8$ \\

& Betweenness $H_0$
& $95.0 \pm 1.9$
& $95.9 \pm 1.4$
& $95.5 \pm 2.4$
& $95.4 \pm 1.6$ \\

& Degree 2 $H_0$
& $96.3 \pm 1.7$
& $96.3 \pm 1.5$
& $96.0 \pm 1.5$
& $96.3 \pm 1.5$ \\

& Nested Ensemble $H_0$
& -- & -- & --
& $\boldsymbol{98.9 \pm 0.9}$ \\

\midrule

\multirow[c]{4}{*}{\textbf{PL}}
& Closeness $H_0$
& $90.4 \pm 2.2$
& $91.6 \pm 1.9$
& $91.1 \pm 1.7$
& $91.4 \pm 1.6$ \\

& Betweenness $H_0$
& $93.1 \pm 2.5$
& $93.6 \pm 2.9$
& $94.1 \pm 2.1$
& $94.1 \pm 2.9$ \\

& Degree 2 $H_0$
& $95.9 \pm 1.5$
& $95.8 \pm 1.3$
& $95.9 \pm 1.4$
& $96.0 \pm 1.7$ \\

& Nested Ensemble $H_0$
& -- & -- & --
& $\boldsymbol{99.1 \pm 1.0}$ \\

\midrule

\multirow[c]{4}{*}{\textbf{PS}}
& Closeness $H_1$
& $99.5 \pm 0.6$
& $99.8 \pm 0.4$
& $99.7 \pm 0.4$
& $99.7 \pm 0.4$ \\

& Betweenness $H_1$
& $99.4 \pm 1.0$
& $100 \pm 0.0$
& $99.8 \pm 0.6$
& $99.7 \pm 0.6$ \\

& Degree 2 $H_1$
& $99.7 \pm 0.4$
& $100 \pm 0.0$
& $100 \pm 0.0$
& $99.9 \pm 0.3$ \\

& Nested Ensemble $H_1$
& -- & -- & --
& $\boldsymbol{100 \pm 0.0}$ \\

\midrule

\multirow[c]{4}{*}{\textbf{BF}}
& Closeness $H_1$
& $98.9 \pm 1.1$
& $99.5 \pm 0.6$
& $99.3 \pm 0.8$
& $99.3 \pm 0.8$ \\

& Betweenness $H_1$
& $99.7 \pm 0.6$
& $99.8 \pm 0.4$
& $99.8 \pm 0.4$
& $99.9 \pm 0.3$ \\

& Degree 2 $H_1$
& $99.5 \pm 0.8$
& $99.2 \pm 1.0$
& $99.5 \pm 0.8$
& $99.7 \pm 0.4$ \\

& Nested Ensemble $H_1$
& -- & -- & --
& $\boldsymbol{100 \pm 0}$ \\

\midrule

\multirow[c]{4}{*}{\textbf{PL}}
& Closeness $H_1$
& $96.2 \pm 1.9$
& $97.9 \pm 1.3$
& $97.7 \pm 0.6$
& $97.9 \pm 1.3$ \\

& Betweenness $H_1$
& $97.8 \pm 1.3$
& $99.3 \pm 0.8$
& $98.4 \pm 0.9$
& $98.8 \pm 1.1$ \\

& Degree 2 $H_1$
& $97.7 \pm 1.5$
& $98.4 \pm 1.5$
& $98.0 \pm 1.5$
& $98.5 \pm 1.1$ \\

& Nested Ensemble $H_1$
& -- & -- & --
& $\boldsymbol{99.9 \pm 0.3}$ \\

\bottomrule
\end{tabularx}
\end{table}

Similarly to what we observed for the Erdős--Rényi model, the proposed nested ensemble learning framework consistently outperforms both individual base models and stacked ensembles, regardless of the summary function (PS, BF, PL) or homology dimension ($H_0$, $H_1$). Across networks generated with different numbers of communities ($k_0 = 2, 5, 10$) the nested ensemble achieved substantial performance gains. For example, in the $H_0$ dimension using Persistence Statistics (PS), the nested ensemble achieved an accuracy of $99.7\%$, representing improvements of up to $2.4\%$ over the best individual base model ($97.3\%$) and $2.5\%$ over the stacked ensembles ($97.2\%$). Similar gains were observed with Betti curves (BF) and Persistence Landscapes (PL).

Performance gains were even more pronounced in the $H_1$ dimension. 
For PS, the nested ensemble achieved a perfect $100\%$ accuracy, matching the best individual base model while exhibiting tighter consistency across folds. 
In the case of BF, the nested ensemble also reached $100\%$, exceeding the top-performing individual and stacked ensembles, which ranged between $99.5\%$ and $99.9\%$. 
Finally, for PL in the $H_1$ dimension, the nested ensemble attained $99.9\%$, yielding improvements of approximately $0.6\%$ to $1.1\%$ over the best base and stacked ensembles (around $98.8\%$).

\subsubsection{Experiment on Simulated Data-Regression Setting}

{This experiment evaluates the effectiveness of the proposed nested ensemble model in a regression task involving synthetic graphs generated using the Watts--Strogatz small-world model. The primary objective is to determine whether topological features extracted via persistent homology offer superior predictive power compared to conventional non-topological summaries of node attributes. We generate a total of 200 graphs, with 150 used for training and 50 for testing. Each graph contains a number of nodes randomly sampled from the interval} $[80, 120]$, and the number of nearest neighbors is fixed at 10. For each graph, the rewiring probability $p$, which serves as the regression target, is sampled uniformly from the range $[0.05, 0.5]$. The learning task is to predict this rewiring probability based solely on the structural characteristics of each graph.

To investigate the impact of different types of feature representations, we compare two~distinct feature extraction strategies. In the first approach, topological features are derived from persistent homology by constructing lower-star filtrations based on node attributes as filter functions. PDs are computed for each filtration and vectorized to obtain fixed-length feature representations. The second approach serves as a non-topological baseline. For each graph, we compute the same node-level attributes—Degree2, Betweenness, and Closeness centrality—that were used as filter functions in the topological case. However, instead of constructing filtrations or computing persistence diagrams, we summarize each attribute using five simple quantile-based statistics: the minimum, first quartile (Q1), median, third quartile (Q3), and maximum. These features are extracted independently for each graph and provide a low-dimensional, interpretable summary of the node attribute distributions, without incorporating any topological or structural information.

For each feature extraction strategy, we train three base models: a neural network model
 (\texttt{nnet}), a bagging model (\texttt{treebag}), and a support vector machine with a radial basis function kernel (\texttt{svmRadial}), all implemented with the default hyperparameter settings provided by the \texttt{caret} package in R~\cite{kuhn2008caret}. Our focus here is not on optimizing the performance of individual models, but rather on maintaining a consistent baseline to assess the relative effectiveness of the different feature extraction strategies. The predictions from these base models are combined using a generalized linear model (GLM) as a Level-1 meta-learner to form the filter-specific ensembles. At Level-2, a Random Forest (RF) is employed to integrate the outputs of the three filter-specific ensembles—$E_{\text{Degree2}}$, $E_{\text{Betweenness}}$, and $E_{\text{Closeness}}$—to produce the final regression prediction. Model performance is evaluated using the Root Mean Squared Error (RMSE), averaged over 10 independent iterations. For the topological feature extraction strategy, we report the performance of the $H_{1}$ features, although $H_{0}$ also demonstrated competitive performance. The regression results for the simulated data under both feature extraction strategies are summarized in {Table~\ref{tab:reddit_binary_combined1}}.

\begin{table}[H]
\centering
\caption{Root Mean Squared Error (RMSE) values for the regression
task on simulated Watts--Strogatz graphs, comparing individual
base models, stacked ensembles, and the proposed nested ensemble.
Results are reported for $H_1$-based topological features and
non-topological graph-theoretic features. Lower RMSE values
indicate better model performance.}
\label{tab:reddit_binary_combined1}

\begin{tabularx}{\textwidth}{@{}lXcccc@{}}
\toprule
\textbf{Summary} &
\textbf{Filter} &
\textbf{nnet} &
\textbf{Bagging} &
\textbf{SvmRadial} &
\textbf{Stacked} \\
\midrule

\multirow[c]{4}{*}{\textbf{PS}}
& Closeness $H_1$
& $3.99 \pm 0.47$
& $5.08 \pm 0.65$
& $4.87 \pm 0.72$
& $4.01 \pm 0.52$ \\

& Betweenness $H_1$
& $4.81 \pm 3.13$
& $5.57 \pm 0.89$
& $5.15 \pm 0.95$
& $4.68 \pm 2.69$ \\

& Degree2 $H_1$
& $8.98 \pm 8.1$
& $5.54 \pm 1.02$
& $4.86 \pm 0.71$
& $6.02 \pm 2.43$ \\

& Nested Ensemble $H_1$
& -- & -- & --
& $\boldsymbol{3.95 \pm 0.57}$ \\

\midrule

\multirow[c]{4}{*}{\textbf{BF}}
& Closeness $H_1$
& $6.85 \pm 0.56$
& $8.50 \pm 0.75$
& $6.86 \pm 0.47$
& $5.56 \pm 0.48$ \\

& Betweenness $H_1$
& $5.62 \pm 0.63$
& $4.56 \pm 0.29$
& $6.57 \pm 0.25$
& $4.28 \pm 0.23$ \\

& Degree2 $H_1$
& $5.21 \pm 0.49$
& $5.62 \pm 0.39$
& $6.68 \pm 0.40$
& $4.76 \pm 0.37$ \\

& Nested Ensemble $H_1$
& -- & -- & --
& $\boldsymbol{4.22 \pm 0.20}$ \\

\midrule

\multicolumn{6}{c}{\textbf{Non-topological Features}} \\
\midrule

& Closeness
& $12.4 \pm 1.07$
& $12.38 \pm 1.66$
& $12.72 \pm 1.20$
& $12.31 \pm 1.11$ \\

& Betweenness
& $12.94 \pm 0.78$
& $12.14 \pm 9.50$
& $11.69 \pm 1.05$
& $11.6 \pm 0.96$ \\

& Degree2
& $6.36 \pm 0.83$
& $6.06 \pm 0.55$
& $6.32 \pm 0.72$
& $6.00 \pm 0.58$ \\

& Nested Ensemble
& -- & -- & --
& $\boldsymbol{5.91 \pm 0.61}$ \\

\bottomrule
\end{tabularx}
\end{table}

Across all experimental settings, the proposed nested ensemble model consistently outperforms both the individual base models and the standard stacked ensembles. When using topological features derived from persistent homology, the nested ensemble achieves the lowest overall RMSE values. Specifically, in the case of the $H_1$ dimension using persistence statistics (PS), the nested ensemble attains an RMSE of $3.95 \pm 0.57$, compared to $4.01 \pm 0.52$ for the best-performing stacked ensemble. A similar trend is observed for $H_1$ using the Betti function (BF), where the nested ensemble achieves an RMSE of $4.22 \pm 0.20$, outperforming the best stacked ensemble at $4.28 \pm 0.23$ and all individual base models.

In contrast, performance deteriorates across all models when using the graph-theoretic approach based on raw filter function summaries. While the nested ensemble still maintains a slight edge over the stacked ensemble ($5.91 \pm 0.61$ vs.\ $6.00 \pm 0.58$), the improvement is less pronounced. Furthermore, individual base models such as the neural network and support vector machine perform significantly worse in this setting, with RMSE values reaching as high as $12.94 \pm 0.78$ and $12.72 \pm 1.20$, respectively. These results suggest that topological features offer a more structured and informative representation for regression tasks than simple summary statistics based on node-level graph attributes.

\subsection{Nested Ensemble on Real Data}
 
\label{ens}
This section evaluates the proposed nested ensemble model on four benchmark datasets from the TUDataset collection: \textit{IMDB-BINARY}, \textit{IMDB-MULTI}, \textit{REDDIT-BINARY}, and \textit{PROTEINS}. These datasets are widely used for benchmarking graph classification algorithms and are derived from real-world networks. Each dataset consists of graphs labeled into distinct classes and exhibits variation in graph size, number of nodes, and class distribution. A detailed summary of their characteristics is presented in Table~\ref{tab:real_data_summary}. Given that some datasets exhibit class imbalance, we apply the ROSE (Random Over-Sampling Examples) technique~\cite{lunardon2014rose, menardi2014training} to balance the training data and mitigate potential bias during model learning. 

\begin{table}[H]
\centering
\caption{Summary of real-world graph datasets used in the
experiments of Section \ref{ens}.}
\label{tab:real_data_summary}

\begin{tabularx}{\textwidth}{@{}Xcccc@{}}
\toprule
\textbf{Dataset} &
\textbf{No. of Graphs} &
\textbf{No. of Classes} &
\textbf{Avg. No. of Nodes} &
\textbf{Avg. No. of Edges} \\
\midrule

IMDB-BINARY
& 1000
& 2
& 19.77
& 96.53 \\

IMDB-MULTI
& 1500
& 3
& 13.00
& 65.94 \\

REDDIT-BINARY
& 2000
& 2
& 429.60
& 497.80 \\

PROTEINS
& 1113
& 2
& 39.31
& 72.82 \\

\bottomrule
\end{tabularx}
\end{table}

For each dataset, a 75--25\% train--test split is performed. PDs are computed for homological dimensions 0 and 1 using lower-star filtrations induced by three filter functions. The filtration is guided by a function $ g $, where $ g $ denotes filter function values normalized to the interval $[0, 1]$. The three filter functions used are (closeness, betweenness, \textit{degree 2}).
 For each set of vectorized features corresponding to each filter function, three base models are trained: Bagging, Random Forest, and Boosting. These base models are combined using a generalized linear model (GLM) as a Level-1 meta-learner, which produces a filter-specific stacked ensemble. A Random Forest (RF) then serves as a Level-2 meta-learner, integrating the predictions from the three filter-specific stacked ensembles to produce the final output of the nested ensemble model. The classification results on the real-world datasets for $H_{0}$-based features are presented in Table~\ref{tab:real_world_combined}. The results clearly show that the nested ensemble learning framework consistently achieves the highest accuracy, outperforming both individual base models and stacked ensembles. {Here, we report results based exclusively on $ H_0 $-based features. Although features} {derived from $ H_1 $ were also competitive, we present the results only for the homological dimension that achieved the superior predictive performance.} Across all topological summary functions, the nested ensemble achieves gains of approximately $0.1\%$ to $5\%$ over the individual base models and $0.1\%$ to $4.9\%$ over the stacked ensembles. This pattern is consistent across all datasets, highlighting the robustness and effectiveness of the nested ensemble framework.

Table~\ref{tab:imdb_reddit_results} {compares} the classification accuracies achieved by the proposed nested ensemble framework against several graph neural network (GNN) and graph kernel baselines on four benchmark datasets. The baseline results are taken directly from their original publications. As shown in the table, the nested ensemble consistently outperforms these baselines across most datasets or ranks among the top three methods when its performance is slightly lower. This highlights the effectiveness of the nested ensemble framework for learning graph-level properties across different domains.
It is important to note, however, that the nested ensemble did not achieve equally strong results across all summary functions, particularly on datasets with imbalanced classes such as PROTEINS. Competitive performance relative to the baselines was observed primarily when using Persistence Statistics (PS), whereas results with Betti curves (BF) and persistence landscapes (PL) were generally less favorable.

\begin{table}[H]
\centering
\caption{Classification accuracy (\%, mean $\pm$ SD) for PS, GPL, and BF across four graph datasets using $H_0$ features.}
\label{tab:real_world_combined}

\begingroup

\begin{tabularx}{\textwidth}{@{}lXcccc@{}}
\toprule
\textbf{Summary} &
\textbf{Filter} &
\textbf{Bagging} &
\textbf{RF} &
\textbf{Boost} &
\textbf{Stacked} \\
\midrule

\multicolumn{6}{c}{\textbf{REDDIT-BINARY}} \\
\midrule
\multirow[c]{4}{*}{\textbf{PS}}
& Closeness $H_0$ & $87.2 \pm 1.5$ & $87.4 \pm 1.1$ & $86.6 \pm 1.5$ & $87.3 \pm 1.5$ \\
& Betweenness $H_0$ & $87.8 \pm 1.9$ & $88.1 \pm 2.0$ & $87.4 \pm 2.1$ & $88.1 \pm 1.7$ \\
& Degree2 $H_0$ & $89.5 \pm 1.0$ & $90.2 \pm 1.3$ & $89.3 \pm 1.5$ & $90.2 \pm 1.3$ \\
& Nested Ensemble $H_0$ & -- & -- & -- & $\boldsymbol{90.3 \pm 1.6}$ \\
\midrule
\multirow[c]{4}{*}{\textbf{GPL}}
& Closeness $H_0$ & $75.7 \pm 1.8$ & $75.5 \pm 1.5$ & $76.8 \pm 1.4$ & $76.9 \pm 1.5$ \\
& Betweenness $H_0$ & $71.7 \pm 1.1$ & $73.9 \pm 3.2$ & $78.6 \pm 1.8$ & $78.8 \pm 1.2$ \\
& Degree2 $H_0$ & $77.7 \pm 1.3$ & $79.4 \pm 2.2$ & $80.3 \pm 1.4$ & $80.3 \pm 1.5$ \\
& Nested Ensemble $H_0$ & -- & -- & -- & $\boldsymbol{83.7 \pm 2.1}$ \\
\midrule
\multirow[c]{4}{*}{\textbf{BF}}
& Closeness $H_0$ & $58.7 \pm 2.0$ & $64.8 \pm 1.2$ & $63.7 \pm 2.3$ & $64.8 \pm 2.2$ \\
& Betweenness $H_0$ & $74.0 \pm 1.5$ & $78.0 \pm 1.1$ & $79.0 \pm 1.2$ & $78.8 \pm 1.5$ \\
& Degree2 $H_0$ & $84.8 \pm 1.5$ & $86.1 \pm 1.4$ & $84.6 \pm 1.7$ & $86.1 \pm 1.4$ \\
& Nested Ensemble $H_0$ & -- & -- & -- & $\boldsymbol{88.1 \pm 1.1}$ \\
\midrule

\multicolumn{6}{c}{\textbf{IMDB-MULTI}} \\
\midrule
\multirow[c]{4}{*}{\textbf{PS}}
& Closeness $H_0$ & $48.7 \pm 2.1$ & $48.5 \pm 2.3$ & $48.0 \pm 2.1$ & $48.7 \pm 2.0$ \\
& Betweenness $H_0$ & $40.4 \pm 1.4$ & $40.6 \pm 0.8$ & $40.2 \pm 1.7$ & $40.9 \pm 1.7$ \\
& Degree2 $H_0$ & $48.7 \pm 3.1$ & $48.8 \pm 2.4$ & $49.4 \pm 2.9$ & $49.4 \pm 2.2$ \\
& Nested Ensemble $H_0$ & -- & -- & -- & $\boldsymbol{54.1 \pm 1.1}$ \\
\midrule
\multirow[c]{4}{*}{\textbf{GPL}}
& Closeness $H_0$ & $47.8 \pm 2.4$ & $47.1 \pm 2.9$ & $46.7 \pm 2.7$ & $47.5 \pm 2.6$ \\
& Betweenness $H_0$ & $39.2 \pm 1.5$ & $39.3 \pm 1.3$ & $39.5 \pm 1.6$ & $40.3 \pm 1.8$ \\
& Degree2 $H_0$ & $46.9 \pm 2.6$ & $47.2 \pm 3.2$ & $47.1 \pm 2.5$ & $47.5 \pm 2.5$ \\
& Nested Ensemble $H_0$ & -- & -- & -- & $\boldsymbol{52.2 \pm 2.8}$ \\
\midrule
\multirow[c]{4}{*}{\textbf{BF}}
& Closeness $H_0$ & $47.8 \pm 2.2$ & $46.7 \pm 3.2$ & $46.4 \pm 1.8$ & $48.2 \pm 2.2$ \\
& Betweenness $H_0$ & $39.1 \pm 1.5$ & $39.2 \pm 1.5$ & $39.9 \pm 2.0$ & $39.3 \pm 1.9$ \\
& Degree2 $H_0$ & $46.9 \pm 1.9$ & $46.5 \pm 2.0$ & $46.3 \pm 2.6$ & $46.9 \pm 1.7$ \\
& Nested Ensemble $H_0$ & -- & -- & -- & $\boldsymbol{52.1 \pm 3.0}$ \\
\midrule

\multicolumn{6}{c}{\textbf{PROTEINS}} \\
\midrule
\multirow[c]{4}{*}{\textbf{PS}}
& Closeness $H_0$ & $63.9 \pm 5.2$ & $69.4 \pm 1.9$ & $63.6 \pm 5.1$ & $71.0 \pm 1.9$ \\
& Betweenness $H_0$ & $55.0 \pm 8.6$ & $68.2 \pm 4.3$ & $64.1 \pm 6.5$ & $70.0 \pm 2.0$ \\
& Degree2 $H_0$ & $64.7 \pm 3.1$ & $66.8 \pm 2.2$ & $64.6 \pm 3.0$ & $68.8 \pm 2.1$ \\
& Nested Ensemble $H_0$ & -- & -- & -- & $\boldsymbol{78.0 \pm 2.3}$ \\
\midrule
\multirow[c]{4}{*}{\textbf{GPL}}
& Closeness $H_0$ & $59.7 \pm 0.2$ & $62.0 \pm 1.4$ & $59.8 \pm 0.5$ & $69.2 \pm 2.4$ \\
& Betweenness $H_0$ & $40.5 \pm 0.2$ & $41.3 \pm 1.7$ & $40.6 \pm 0.3$ & $51.8 \pm 6.2$ \\
& Degree2 $H_0$ & $42.4 \pm 5.9$ & $61.7 \pm 7.0$ & $40.5 \pm 4.0$ & $61.7 \pm 6.1$ \\
& Nested Ensemble $H_0$ & -- & -- & -- & $\boldsymbol{71.8 \pm 2.9}$ \\
\midrule
\multirow[c]{4}{*}{\textbf{BF}}
& Closeness $H_0$ & $46.5 \pm 2.8$ & $51.7 \pm 2.8$ & $53.5 \pm 3.4$ & $54.4 \pm 3.2$ \\
& Betweenness $H_0$ & $49.3 \pm 2.4$ & $52.5 \pm 1.2$ & $53.7 \pm 1.7$ & $55.1 \pm 1.9$ \\
& Degree2 $H_0$ & $48.9 \pm 2.6$ & $51.1 \pm 1.6$ & $51.9 \pm 2.0$ & $53.8 \pm 2.4$ \\
& Nested Ensemble $H_0$ & -- & -- & -- & $\boldsymbol{68.0 \pm 3.3}$ \\
\midrule

\multicolumn{6}{c}{\textbf{IMDB-BINARY}} \\
\midrule
\multirow[c]{4}{*}{\textbf{PS}}
& Closeness $H_0$ & $70.1 \pm 2.7$ & $70.1 \pm 2.6$ & $67.6 \pm 2.4$ & $70.9 \pm 2.9$ \\
& Betweenness $H_0$ & $64.3 \pm 2.3$ & $64.5 \pm 1.7$ & $64.2 \pm 2.0$ & $65.0 \pm 2.9$ \\
& Degree2 $H_0$ & $71.3 \pm 2.3$ & $70.7 \pm 2.0$ & $70.8 \pm 1.6$ & $71.0 \pm 1.6$ \\
& Nested Ensemble $H_0$ & -- & -- & -- & $\boldsymbol{72.1 \pm 2.6}$ \\
\midrule
\multirow[c]{4}{*}{\textbf{GPL}}
& Closeness $H_0$ & $68.3 \pm 2.4$ & $69.6 \pm 2.7$ & $66.7 \pm 2.2$ & $69.7 \pm 2.6$ \\
& Betweenness $H_0$ & $63.2 \pm 2.8$ & $61.2 \pm 1.9$ & $62.0 \pm 2.1$ & $62.4 \pm 2.7$ \\
& Degree2 $H_0$ & $68.9 \pm 2.3$ & $68.5 \pm 3.6$ & $69.4 \pm 3.1$ & $69.9 \pm 2.9$ \\
& Nested Ensemble $H_0$ & -- & -- & -- & $\boldsymbol{78.2 \pm 1.7}$ \\
\midrule
\multirow[c]{4}{*}{\textbf{BF}}
& Closeness $H_0$ & $68.4 \pm 1.9$ & $67.3 \pm 0.3$ & $63.2 \pm 2.1$ & $68.8 \pm 1.9$ \\
& Betweenness $H_0$ & $62.8 \pm 2.7$ & $63.0 \pm 2.5$ & $61.8 \pm 3.4$ & $63.6 \pm 3.0$ \\
& Degree2 $H_0$ & $69.3 \pm 2.8$ & $70.0 \pm 2.7$ & $69.4 \pm 2.6$ & $70.8 \pm 2.6$ \\
& Nested Ensemble $H_0$ & -- & -- & -- & $\boldsymbol{79.0 \pm 2.3}$ \\

\bottomrule
\end{tabularx}
\endgroup
\end{table}

\begin{table}[H]
\centering
\caption{Graph classification accuracy (\%) across benchmark datasets. The top section reports results from baseline methods, taken directly from their original publications. The bottom section presents results from the proposed nested ensemble learning framework using three summary functions (PS, BF, PL), evaluated on $H_0$-based features.}
\label{tab:imdb_reddit_results}
\begin{tabularx}{\textwidth}{@{}Xcccc@{}}
\toprule
\textbf{Model} &
\textbf{IMDB-BINARY} &
\textbf{REDDIT-BINARY} &
\textbf{IMDB-MULTI} &
\textbf{PROTEINS} \\
\midrule

\multicolumn{5}{l}{\textbf{Previous Methods}} \\
\midrule

2-WL-GNN~\cite{damke2020novelhigherorderweisfeilerlehmangraph}
& $72.2 \pm 3.1$ & $89.4 \pm 2.6$ & -- & $76.2 \pm 3.3$ \\

GIN-0~\cite{xu2019powerfulgraphneuralnetworks}
& $75.1 \pm 5.1$ & $\boldsymbol{92.4 \pm 2.5}$ & $52.3 \pm 2.8$ & $76.2 \pm 2.8$ \\

GraphSAGE~\cite{errica2022faircomparisongraphneural}
& $68.8 \pm 4.5$ & $84.3 \pm 1.9$ & $47.6 \pm 3.5$ & $75.9 \pm 3.2$ \\

DGCNN~\cite{zhang2018end}
& $69.2 \pm 3.0$ & $87.8 \pm 2.5$ & $45.6 \pm 3.4$ & $72.9 \pm 3.5$ \\

AWE~\cite{ivanov2018anonymous}
& $74.5 \pm 5.8$ & $87.9 \pm 2.5$ & $51.1 \pm 3.6$ & -- \\

MP-HSM-C~\cite{kerber2024graphcodelearningmultiparameterpersistent}
& $74.8 \pm 2.5$ & -- & -- & $74.6 \pm 2.1$ \\

\midrule
\multicolumn{5}{l}{\textbf{Ours: Nested Ensemble Methods}} \\
\midrule

Nested Ensemble (PS) $H_0$
& $72.1 \pm 2.6$
& $90.3 \pm 1.6$
& $\boldsymbol{54.1 \pm 1.1}$
& $\boldsymbol{78.0 \pm 2.3}$ \\

Nested Ensemble (BF) $H_0$
& $\boldsymbol{79.0 \pm 2.3}$
& $88.1 \pm 1.1$
& $52.1 \pm 3.0$
& $68.0 \pm 3.3$ \\

Nested Ensemble (PL) $H_0$
& $78.2 \pm 1.7$
& $83.7 \pm 2.1$
& $52.2 \pm 2.8$
& $71.8 \pm 2.9$ \\

\bottomrule
\end{tabularx}
\end{table}

\section{Conclusions}
\label{sec:conclusion}

We propose an efficient framework for extracting topological features directly from graph structures, without the need to embed them into metric spaces. Our method builds simplicial complexes from graphs and computes persistent homology using lower-star filtrations induced by three filter functions: closeness centrality, betweenness centrality, and a custom second-order degree centrality degree2. This approach captures the intrinsic topology of the graph while maintaining computational efficiency.

Building on these topological features, we propose a \emph{nested ensemble learning} framework that leverages these features to build powerful model for both classification and regression tasks on graph-structured data. This framework extends standard stacking approaches by incorporating multiple filter functions, selected in a data-driven manner. This design allows the model to learn from diverse topological perspectives, thereby enhancing both accuracy and robustness. For each filter function, we train three base models whose predictions are combined through a meta-learner to create a filter-specific ensemble. A second-level meta-learner then learns to optimally combine these ensembles, effectively assigning weights to each filter function based on the data. We extensively evaluated this framework on both synthetic and real-world graph datasets across classification and regression tasks. In all experiments, our method consistently demonstrated performance gains. On the real-world datasets, it generally outperformed state-of-the-art baseline methods, and even in the worst cases still ranked among the top three models.

Despite its strong performance, a key limitation of the proposed framework is its sensitivity to class imbalance. When one class is significantly underrepresented, the nested ensemble may underperform relative to individual models, indicating a bias toward majority class patterns. As part of future work, we plan to enhance the framework’s ability to handle imbalanced data more effectively. Additionally, we aim to explore the theoretical underpinnings of the architecture, including a formal analysis of its bias--variance trade-off and generalization properties. Such insights would help clarify when and why the nested ensemble performs well and further support its application in real-world settings.

\section*{Data Availability}
All data supporting the findings of this study are included in this article. The benchmark datasets are publicly available at \emph{TUDatasets}
~\cite{kersting2016benchmark}. Further inquiries may be directed to the corresponding author.

\bibliographystyle{unsrtnat}
\bibliography{reference}

@article{erdos1959random,
  author    = {Erdős, P. and Rényi, A.},
  title     = {On random graphs},
  journal   = {Publicationes Mathematicae},
  volume    = {6},
  pages     = {290--297},
  year      = {1959}
}

@article{berry2020functional,
  title={Functional summaries of persistence diagrams},
  author={Berry, Eric and Chen, Yen-Chi and Cisewski-Kehe, Jessi and Fasy, Brittany T.},
  journal={Journal of the American Statistical Association},
  year={2020},
  publisher={Taylor \& Francis}
}

@article{ali2023survey,
  author    = {Ali, Dalia and Asaad, Ahmad and Jimenez, Maria J. and Nanda, Vinesh and Paluzo-Hidalgo, Elena and Soriano-Trigueros, Mario},
  title     = {A Survey of Vectorization Methods in Topological Data Analysis},
  journal   = {IEEE Transactions on Pattern Analysis and Machine Intelligence},
  year      = {2023},
  publisher = {IEEE}
}

@article{zhang2020topological,
  title        = {Topological portraits of multiscale coordination dynamics},
  author       = {Zhang, Mengsen and Kalies, William D. and Kelso, J.A. Scott and Tognoli, Emmanuelle},
  journal      = {Journal of Neuroscience Methods},
  volume       = {339},
  pages        = {108672},
  year         = {2020},
  publisher    = {Elsevier},
  note         = {\href{https://doi.org/10.1016/j.jneumeth.2020.108672}{[CrossRef]}}
}

@book{edelsbrunner2010computational,
  title={Computational Topology: An Introduction},
  author={Edelsbrunner, Herbert and Harer, John L},
  year={2010},
  publisher={American Mathematical Society}
}

@article{ghrist2008barcodes,
  title={Barcodes: The persistent topology of data},
  author={Ghrist, Robert},
  journal={Bulletin of the American Mathematical Society},
  volume={45},
  number={1},
  pages={61--75},
  year={2008}
}

@article{otter2017roadmap,
  title={A roadmap for the computation of persistent homology},
  author={Otter, Nina and Porter, Mason A and Tillmann, Ulrike and Grindrod, Peter and Harrington, Heather A},
  journal={EPJ Data Science},
  volume={6},
  number={1},
  pages={1--38},
  year={2017},
  publisher={Springer}
}

@article{bubenik2015statistical,
  title={Statistical topological data analysis using persistence landscapes},
  author={Bubenik, Peter},
  journal={Journal of Machine Learning Research},
  volume={16},
  number={1},
  pages={77--102},
  year={2015}
}

@article{chazal2021introduction,
  title        = {An introduction to topological data analysis: fundamental and practical aspects for data scientists},
  author       = {Chazal, Fr{\'e}d{\'e}ric and Michel, Bertrand},
  journal      = {Frontiers in Artificial Intelligence},
  volume       = {4},
  pages        = {667963},
  year         = {2021},
  publisher    = {Frontiers Media SA},
  note         = {\href{https://doi.org/10.3389/frai.2021.667963}{[CrossRef]}}
}

@article{chung2022persistence,
  title        = {Persistence curves: A canonical framework for summarizing persistence diagrams},
  author       = {Chung, Yu-Min and Lawson, Austin},
  journal      = {Advances in Computational Mathematics},
  volume       = {48},
  number       = {1},
  pages        = {1--42},
  year         = {2022},
  publisher    = {Springer},
  note         = {\href{https://doi.org/10.1007/s10444-021-09905-3}{[CrossRef]}}
}

@inproceedings{chazal2014stochastic,
  title={Stochastic convergence of persistence landscapes and silhouettes},
  author={Chazal, Fr{\'e}d{\'e}ric and Fasy, Brittany Terese and Lecci, Fabrizio and Rinaldo, Alessandro and Wasserman, Larry},
  booktitle={Proceedings of the Thirtieth Annual Symposium on Computational Geometry},
  pages={474--483},
  year={2014},
  organization={ACM}
}

@article{islambekov2024vector,
  title={Vector Summaries of Persistence Diagrams for Permutation-based Hypothesis Testing},
  author={Islambekov, Umar D. and Pathirana, Heshan},
  journal={Foundations of Data Science},
  volume={6},
  number={1},
  pages={41--61},
  year={2024}
}

@article{watts1998smallworld,
  title        = {Collective dynamics of ‘small-world’ networks},
  author       = {Watts, Duncan J. and Strogatz, Steven H.},
  journal      = {Nature},
  volume       = {393},
  number       = {6684},
  pages        = {440--442},
  year         = {1998},
  note         = {\href{https://doi.org/10.1038/30918}{[CrossRef]}}
}

@manual{TDAvec2025,
  title        = {{TDAvec}: Vector Summaries of Persistence Diagrams},
  author       = {Umar Islambekov and Aleksei Luchinsky},
  year         = {2025},
  note         = {\href{https://doi.org/10.32614/CRAN.package.TDAvec}{[CrossRef]}},
  url          = {https://CRAN.R-project.org/package=TDAvec}
}

@misc{kersting2016benchmark,
  author       = {Kersting, Kristian and Kriege, Nils M. and Morris, Christopher and Mutzel, Petra and Neumann, Marc},
  title        = {Benchmark data sets for graph kernels},
  year         = {2016}
}

@article{wolpert1992stacked,
  title   = {Stacked Generalization},
  author  = {Wolpert, David H.},
  journal = {Neural Networks},
  volume  = {5},
  number  = {2},
  pages   = {241--259},
  year    = {1992},
  doi     = {10.1016/S0893-6080(05)80023-1}
}

@article{carlsson2009topology,
  title   = {Topology and Data},
  author  = {Carlsson, Gunnar},
  journal = {Bulletin of the American Mathematical Society},
  volume  = {46},
  number  = {2},
  pages   = {255--308},
  year    = {2009},
  doi     = {10.1090/S0273-0979-09-01249-X}
}

@article{carriere2020perslay,
  title={PersLay: A neural network layer for persistence diagrams and new graph topological signatures},
  author={Carri{\`e}re, Mathieu and Cuturi, Marco and Oudot, Steve},
  journal={Proceedings of the 23rd International Conference on Artificial Intelligence and Statistics (AISTATS)},
  volume={108},
  pages={2786--2796},
  year={2020}
}

@misc{islambekov2023fasttopologicalapproachpredicting,
  title        = {A fast topological approach for predicting anomalies in time-varying graphs},
  author       = {Umar Islambekov and Hasani Pathirana and Omid Khormali and Cuneyt Akcora and Ekaterina Smirnova},
  year         = {2023},
  note         = {\href{https://arxiv.org/abs/2305.06523}{[CrossRef]}}
}

@misc{zhao2019learningmetricspersistencebasedsummaries,
  title        = {Learning metrics for persistence-based summaries and applications for graph classification},
  author       = {Qi Zhao and Yusu Wang},
  year         = {2019},
  note         = {\href{https://arxiv.org/abs/1904.12189}{[CrossRef]}}
}

@misc{damke2020novelhigherorderweisfeilerlehmangraph,
  title        = {A Novel Higher-order Weisfeiler-Lehman Graph Convolution},
  author       = {Clemens Damke and Vitalik Melnikov and Eyke Hüllermeier},
  year         = {2020},
  note         = {\href{https://arxiv.org/abs/2007.00346}{[CrossRef]}}
}

@misc{xu2019powerfulgraphneuralnetworks,
  title        = {How Powerful are Graph Neural Networks?},
  author       = {Keyulu Xu and Weihua Hu and Jure Leskovec and Stefanie Jegelka},
  year         = {2019},
  note         = {\href{https://arxiv.org/abs/1810.00826}{[CrossRef]}}
}

@misc{errica2022faircomparisongraphneural,
  title        = {A Fair Comparison of Graph Neural Networks for Graph Classification},
  author       = {Federico Errica and Marco Podda and Davide Bacciu and Alessio Micheli},
  year         = {2022},
  note         = {\href{https://arxiv.org/abs/1912.09893}{[CrossRef]}}
}

@misc{hajij2021persistenthomologygraphsrepresentation,
  title        = {Persistent Homology and Graphs Representation Learning},
  author       = {Mustafa Hajij and Ghada Zamzmi and Xuanting Cai},
  year         = {2021},
  note         = {\href{https://arxiv.org/abs/2102.12926}{[CrossRef]}}
}

@article{bu2023interplay,
  title        = {Interplay between Topology and Edge Weights in Real-World Graphs: Concepts, Patterns, and an Algorithm},
  author       = {Bu, Fanchen and Kang, Shinhwan and Shin, Kijung},
  journal      = {arXiv preprint arXiv:2305.09083},
  year         = {2023},
  note         = {\href{https://arxiv.org/abs/2305.09083}{[CrossRef]}}
}

@misc{hofer2018deeplearningtopologicalsignatures,
  title        = {Deep Learning with Topological Signatures},
  author       = {Christoph Hofer and Roland Kwitt and Marc Niethammer and Andreas Uhl},
  year         = {2018},
  note         = {\href{https://arxiv.org/abs/1707.04041}{[CrossRef]}}
}

@inproceedings{obray2021filtration,
  title        = {Filtration Curves for Graph Representation},
  author       = {O'Bray, Leslie and Rieck, Bastian and Borgwardt, Karsten},
  booktitle    = {Proceedings of the 27th ACM SIGKDD Conference on Knowledge Discovery and Data Mining},
  pages        = {1304--1314},
  year         = {2021},
  organization = {ACM},
  note         = {\href{https://doi.org/10.1145/3447548.3467352}{[CrossRef]}}
}

@article{agerberg2025certifying,
  title        = {Certifying Robustness via Topological Representations},
  author       = {Agerberg, Jens and Guidolin, Andrea and Martinelli, Andrea and Roos Hoefgeest, Pepijn and Eklund, David and Scolamiero, Martina},
  journal      = {arXiv preprint arXiv:2501.10876},
  year         = {2025},
  note         = {\href{https://arxiv.org/abs/2501.10876}{[CrossRef]}}
}

@inproceedings{ivanov2018anonymous,
  title={Anonymous Walk Embeddings},
  author={Ivanov, Sergey and Burnaev, Evgeny},
  booktitle={Proceedings of the 35th International Conference on Machine Learning (ICML)},
  pages={2186--2195},
  year={2018}
}

@inproceedings{zhang2018end,
  title={An End-to-End Deep Learning Architecture for Graph Classification},
  author={Zhang, Muhan and Cui, Zhicheng and Neumann, Marion and Chen, Yixin},
  booktitle={Proceedings of the Thirty-Second AAAI Conference on Artificial Intelligence (AAAI-18)},
  year={2018}
}

@article{Holland1983,
  title        = {Stochastic blockmodels: First steps},
  author       = {Paul W. Holland and Kathryn Blackmond Laskey and Samuel Leinhardt},
  journal      = {Social Networks},
  volume       = {5},
  number       = {2},
  pages        = {109--137},
  year         = {1983},
  publisher    = {Elsevier},
  note         = {\href{https://doi.org/10.1016/0378-8733(83)90021-7}{[CrossRef]}}
}

@inproceedings{nishikawa2023adaptive,
  title        = {Adaptive Topological Feature via Persistent Homology: Filtration Learning for Point Clouds},
  author       = {Naoki Nishikawa and Yuichi Ike and Kenji Yamanishi},
  booktitle    = {Advances in Neural Information Processing Systems 36 (NeurIPS 2023)},
  year         = {2023},
  note         = {\href{https://proceedings.neurips.cc/paper_files/paper/2023/file/1d49235669869ab737c1da9d64b7c769-Paper-Conference.pdf}{[CrossRef]}}
}

@misc{cornell2021usingtopologicalautoencodersfiltering,
  title        = {Using topological autoencoders as a filtering function for global and local topology},
  author       = {Filip Cornell},
  year         = {2021},
  note         = {\href{https://arxiv.org/abs/2012.03383}{[CrossRef]}}
}

@misc{horn2022topologicalgraphneuralnetworks,
  title        = {Topological Graph Neural Networks},
  author       = {Max Horn and Edward De Brouwer and Michael Moor and Yves Moreau and Bastian Rieck and Karsten Borgwardt},
  year         = {2022},
  note         = {\href{https://arxiv.org/abs/2102.07835}{[CrossRef]}}
}

@article{lunardon2014rose,
  author    = {Lunardon, Nicola and Menardi, Giovanna and Torelli, Nicola},
  title     = {{ROSE}: A Package for Binary Imbalanced Learning},
  journal   = {The R Journal},
  volume    = {6},
  pages     = {82--92},
  year      = {2014}
}

@article{menardi2014training,
  author    = {Menardi, Giovanna and Torelli, Nicola},
  title     = {Training and assessing classification rules with imbalanced data},
  journal   = {Data Mining and Knowledge Discovery},
  volume    = {28},
  pages     = {92--122},
  year      = {2014}
}

@Article{kuhn2008caret,
  author  = {Max Kuhn},
  title   = {Building Predictive Models in R Using the caret Package},
  journal = {Journal of Statistical Software},
  year    = {2008},
  volume  = {28},
  number  = {5},
  pages   = {1--26},
  note   = {\href{https://doi.org/10.18637/jss.v028.i05}{[CrossRef]}}
}

@misc{wikipedia:reduced_homology,
  title        = {Reduced homology},
  author       = {{Wikipedia contributors}},
  year         = {2026},
  howpublished = {\url{https://en.wikipedia.org/wiki/Reduced_homology}},
  note         = {Accessed: November 10, 2026}
}

@article{cohen2007stability,
  title     = {Stability of persistence diagrams},
  author    = {Cohen-Steiner, David and Edelsbrunner, Herbert and Harer, John},
  journal   = {Discrete \& Computational Geometry},
  volume    = {37},
  number    = {1},
  pages     = {103--120},
  year      = {2007},
  publisher = {Springer}
}

@misc{kerber2024graphcodelearningmultiparameterpersistent,
      title={Graphcode: Learning from multiparameter persistent homology using graph neural networks}, 
      author={Michael Kerber and Florian Russold},
      year={2024},
      eprint={2405.14302},
      archivePrefix={arXiv},
      primaryClass={math.AT},
note   = {\href{https://arxiv.org/abs/2405.14302}{[CrossRef]}}
}

@inproceedings{NEURIPS2023_d75c474b,
 author = {Loiseaux, David and Scoccola, Luis and Carri\`{e}re, Mathieu and Botnan, Magnus Bakke and OUDOT, Steve},
 booktitle = {Advances in Neural Information Processing Systems},
 editor = {A. Oh and T. Naumann and A. Globerson and K. Saenko and M. Hardt and S. Levine},
 pages = {68316--68342},
 publisher = {Curran Associates, Inc.},
 title = {Stable Vectorization of Multiparameter Persistent Homology using Signed Barcodes as Measures},
note   ={\url{https://proceedings.neurips.cc/paper_files/paper/2023/file/d75c474bc01735929a1fab5d0de3b189-Paper-Conference.pdf} [CrossRef]},
 volume = {36},
 year = {2023}
}

@article{hajij2017persistent,
  title={Persistent homology guided exploration of time-varying graphs},
  author={Hajij, Mustafa and Wang, Bei and Scheidegger, Carlos and Rosen, Paul},
  journal={arXiv preprint arXiv:1707.06683},
  year={2017}
}

@misc{you2022comparingmultiplelatentspace,
      title={Comparing multiple latent space embeddings using topological analysis}, 
      author={Kisung You and Ilmun Kim and Ick Hoon Jin and Minjeong Jeon and Dennis Shung},
      year={2022},
      eprint={2208.12435},
      archivePrefix={arXiv},
      primaryClass={stat.ME},
note   = {\href{https://arxiv.org/abs/2208.12435}{[CrossRef]}}
}

@misc{adams2016persistenceimagesstablevector,
      title={Persistence Images: A Stable Vector Representation of Persistent Homology}, 
      author={Henry Adams and Sofya Chepushtanova and Tegan Emerson and Eric Hanson and Michael Kirby and Francis Motta and Rachel Neville and Chris Peterson and Patrick Shipman and Lori Ziegelmeier},
      year={2016},
      eprint={1507.06217},
      archivePrefix={arXiv},
      primaryClass={cs.CG},
note   = {\href{https://arxiv.org/abs/1507.06217}{[CrossRef]}} 
}

@inbook{Bubenik_2020,
   title={The Persistence Landscape and Some of Its Properties},
   ISBN={9783030434083},
   ISSN={2197-8549},
   booktitle={Topological Data Analysis},
   publisher={Springer International Publishing},
   author={Bubenik, Peter},
   year={2020},
   pages={97–117} }

@misc{chazal2013stochasticconvergencepersistencelandscapes,
      title={Stochastic Convergence of Persistence Landscapes and Silhouettes}, 
      author={Frédéric Chazal and Brittany Terese Fasy and Fabrizio Lecci and Alessandro Rinaldo and Larry Wasserman},
      year={2013},
      eprint={1312.0308},
      archivePrefix={arXiv},
      primaryClass={math.ST},
}

@article{atienza2020persistententropy,
  title={On the stability of persistent entropy and new summary functions for topological data analysis},
  author={Atienza, N. and Gonzalez-D{\'\i}az, R. and Soriano-Trigueros, M.},
  journal={Pattern Recognition},
  volume={107},
  pages={107509},
  year={2020},
  publisher={Elsevier}
}

@misc{niepert2016learningconvolutionalneuralnetworks,
      title={Learning Convolutional Neural Networks for Graphs}, 
      author={Mathias Niepert and Mohamed Ahmed and Konstantin Kutzkov},
      year={2016},
      eprint={1605.05273},
      archivePrefix={arXiv},
      primaryClass={cs.LG},
}

@misc{johnson2021instabilitybettisequencepersistent,
      title={Instability of the Betti Sequence for Persistent Homology and a Stabilized Version of the Betti Sequence}, 
      author={Megan Johnson and Jae-Hun Jung},
      year={2021},
      eprint={2109.09218},
      archivePrefix={arXiv},
      primaryClass={math.AT},
note   = {\href{https://arxiv.org/abs/2109.09218}{[CrossRef]}} 
}

\end{document}